\documentclass[10pt,journal,compsoc]{IEEEtran}
\usepackage[cmex10]{amsmath}
\usepackage{algorithmic}
\usepackage{algorithm}
\usepackage{array}
\usepackage{textcomp}
\usepackage{stfloats}
\usepackage{url}
\usepackage{verbatim}
\usepackage{graphicx}
\usepackage{hyperref}
\usepackage{makecell}
\usepackage{color}
\usepackage{xcolor}
\usepackage{bbm}
\usepackage{booktabs}
\usepackage{multirow}
\usepackage{bm}
\usepackage{colortbl}
\usepackage{amssymb}
\usepackage{enumitem}
\usepackage{cleveref}
\usepackage{dutchcal}

\crefname{figure}{Fig.}{Fig.}  
\crefname{table}{Table}{Tables} 
\crefname{equation}{Eq.}{Eqs.}
\crefname{section}{Section}{Section}

\ifCLASSOPTIONcompsoc
  \usepackage[nocompress]{cite}
\else
  \usepackage{cite}
\fi

\ifCLASSINFOpdf
\else
\fi
\ifCLASSOPTIONcompsoc
 \usepackage[caption=false,font=footnotesize,labelfont=sf,textfont=sf]{subfig}
\else
 \usepackage[caption=false,font=footnotesize]{subfig}
\fi

\begin{document}
%
\title{Generalizing Vision-Language Models to Novel Domains: A Comprehensive Survey}
%
%
%
%

\author{Xinyao Li,
        Jingjing Li,
        Fengling Li,
        Lei Zhu,
        Yang Yang,
        and Heng Tao Shen,~\IEEEmembership{Fellow,~IEEE}
\IEEEcompsocitemizethanks{
\IEEEcompsocthanksitem Xinyao Li, Jingjing Li, Yang Yang and Heng Tao Shen are with University of Electronic Science and Technology of China, Chengdu 610054, China.
\IEEEcompsocthanksitem Fengling Li is with University of Technology Sydney. Lei Zhu is with Tongji University, Shanghai 200070, China.}
\thanks{Manuscript received ; revised .}}

%
%

\markboth{Journal of \LaTeX\ Class Files,~Vol.~13, No.~9, September~2014}%
{Shell \MakeLowercase{\textit{et al.}}: Bare Demo of IEEEtran.cls for Computer Society Journals}
%

\IEEEtitleabstractindextext{%
\begin{abstract}
Recently, vision-language pretraining has emerged as a transformative technique that integrates the strengths of both visual and textual modalities, resulting in powerful vision-language models (VLMs). Leveraging web-scale pretraining data, these models exhibit strong zero-shot capabilities. However, their performance often deteriorates when confronted with domain-specific or specialized generalization tasks. To address this, a growing body of research focuses on transferring or generalizing the rich knowledge embedded in VLMs to various downstream applications. This survey aims to comprehensively summarize the generalization settings, methodologies, benchmarking and results in VLM literatures. Delving into the typical VLM structures, current literatures are categorized into \textit{prompt-based}, \textit{parameter-based} and \textit{feature-based} methods according to the transferred modules. The  differences and characteristics in each category are furthered summarized and discussed by revisiting the typical \textit{transfer learning (TL) settings}, providing novel interpretations for TL in the era of VLMs. Popular benchmarks for VLM generalization are further introduced with thorough performance comparisons among the reviewed methods. Following the advances in large-scale generalizable pretraining, this survey also discusses the relations and differences between VLMs and up-to-date multimodal large language models (MLLM), e.g., DeepSeek-VL. By systematically reviewing the surging literatures in vision-language research from a novel and practical generalization prospective, this survey contributes to a clear landscape of current and future multimodal researches.
\end{abstract}

}

\maketitle

\IEEEdisplaynontitleabstractindextext

\IEEEraisesectionheading{\section{Introduction}\label{sec:introduction}}

\IEEEPARstart{D}{eep} neural networks have achieved remarkable success across a wide range of practical applications. Taking vision models as an example, the progression from AlexNet~\cite{krizhevsky2012imagenet} to ResNet~\cite{he2016deep} and Vision Transformers~\cite{dosovitskiy2020image} has significantly advanced both model scale and representational power. However, training such large-scale models effectively demands substantial labeled data and computational resources. To address this, the concept of \textit{foundation models} has emerged—models pretrained on massive datasets to acquire general-purpose knowledge, which can then be transferred to various downstream tasks~\cite{zhuang2020comprehensive}. For instance, the ResNet family pretrained on ImageNet~\cite{deng2009imagenet} has served as a cornerstone in a wide array of vision tasks, including classification~\cite{he2016deep} and object recognition~\cite{zou2023object}. Parallel advancements have occurred in natural language processing, with the development of influential models such as the Transformer~\cite{vaswani2017attention}, BERT~\cite{devlin2019bert}, GPT-2~\cite{radford2019language}, and GPT-3~\cite{brown2020language}. While these models excel in their respective single-modality domains, they inherently lack the ability to perceive and reason over multimodal information.

\begin{figure}[!t]
    \centering
    \includegraphics[width=0.5\textwidth]{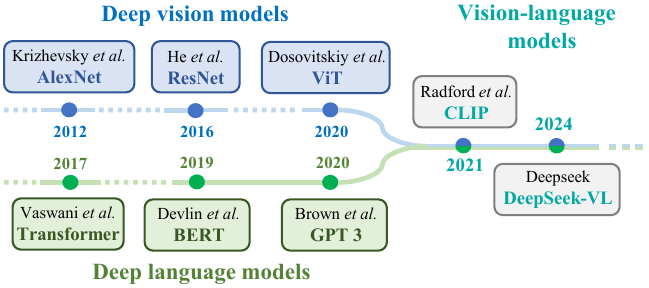}
    \vspace{-8pt}
    \caption{Examples of vision, language and vision-language deep models.}
    \label{fig:intro}
    \vspace{-8pt}
\end{figure}

As shown in \cref{fig:intro}, the emergence of the contrastive language-image pretraining paradigm~\cite{clip} has fundamentally reshaped the landscape of vision-language learning. Leveraging 400M web-crawled image-text pairs, Radford \textit{et al.} introduced the first vision-language foundation model, CLIP~\cite{clip}, which learns by pulling together semantically aligned image-text pairs and pushing apart mismatched ones. This large-scale contrastive pretraining equips CLIP with impressive zero-shot capabilities across a variety of tasks, including image classification~\cite{clip}, object detection~\cite{zhong2022regionclip}, and video-text retrieval~\cite{luo2022clip4clip}. Subsequent works have further extended the capabilities of VLMs by enlarging and denoising pretraining datasets~\cite{jia2021scaling,yu2022coca,li2022blip}, exploring diverse pretraining strategies~\cite{dai2023instructblipgeneralpurposevisionlanguagemodels,liu2023visual}, and incorporating multilingual data~\cite{chen2022pali,chen2023pali,nguyen2024multilingual}.

Despite their outstanding performance on general tasks, generalizing their pretrained knowledge to more specialized downstream tasks is non-trivial. Without proper transfer, pretrained VLMs struggle in handling out-of-distribution (OOD) data like satellite images~\cite{eurosat} and diverse fine-grained images~\cite{flowers,aircraft}. While the typical pretraining - fine-tuning paradigm for knowledge transfer is still applicable, appropriate full-tuning for VLMs has proved tricky. Direct tuning with insufficient target data may disturb the aligned vision-language representations and lead to degraded performances~\cite{kumar2022fine,gao2024clip,zhang2022tip}. Therefore, how to elegantly generalize the pretrained knowledge in VLMs to downstream tasks with minimum computation and annotation expenses has become a research hotspot. Noticing VLMs' multimodal nature, researchers have dedicated to adapt well-developed single-modality transfer methods, e.g., prompt tuning~\cite{lester2021power}, adapters~\cite{houlsby2019parameter}, distillation~\cite{gou2021knowledge}, to VLMs~\cite{coop,cocoop,gao2024clip,li2023distilling}. Powered by vast general knowledge, VLMs are becoming task-agnostic solvers by setting new strong baselines on various transfer learning (TL) scenarios with their strong zero-shot abilities, e.g., unsupervised domain adaptation (UDA)~\cite{daprompt,unimos,damp}, domain generalization (DG)~\cite{bai2024soft,addepalli2024leveraging,chen2024practicaldg}, test-time adaptation (TTA)~\cite{tpt,tda,difftpt}, etc. With such universal success, we ask: \textit{How is knowledge transfer different in the era of VLMs?}

\begin{figure*}[ht]
    \centering
    \includegraphics[width=0.9\textwidth]{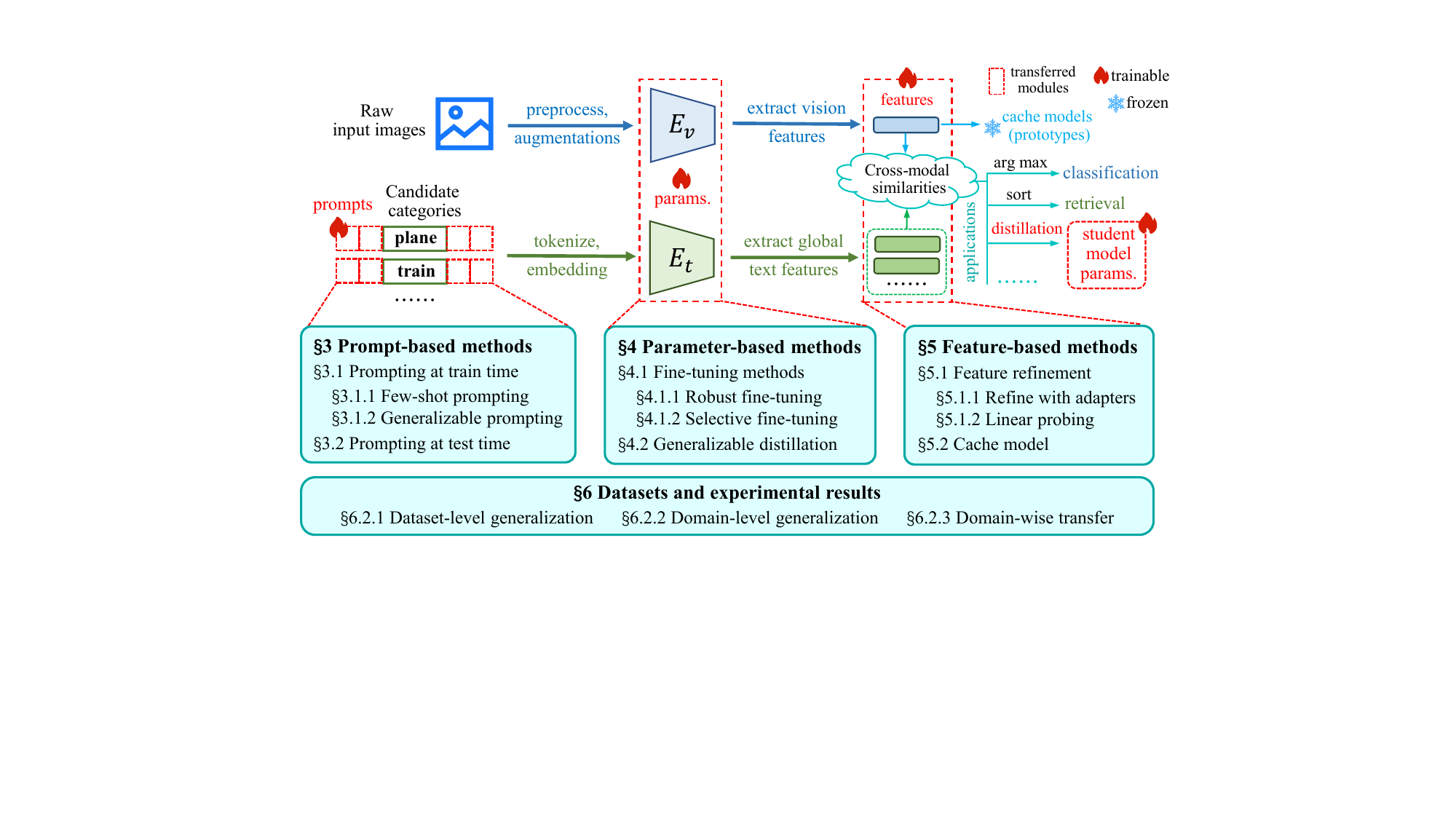}
    \vspace{-8pt}
    \caption{This survey is organized based on the typical and most investigated dual-branch VLM structure, where generalization methods are categorized according to the transferred modules in the VLM.}
    \label{fig:framework}
    \vspace{-8pt}
\end{figure*}

In response to the question, this work conducts a comprehensive literature review on the generalization of pretrained VLMs. Existing surveys~\cite{zhang2024vision,du2022survey,ghosh2024exploring} mainly focus on the \textit{pretraining} of VLMs, including model structures, pretraining objectives and datasets. While survey~\cite{zhang2024vision} mentions TL for VLMs, only a part of few-shot adaptation methods are included, and the differences in transfer setting are not considered. This paper presents the first survey that concentrates on the knowledge \textit{transfer and generalization} of VLMs. As shown in \cref{fig:framework}, our analysis base on the most researched dual-branch VLM, e.g., CLIP~\cite{clip}. We identify and categorize the key components in the VLM and summarize corresponding transfer methods. (1) Prompt-based methods only tunes the textual prompt embeddings to control the VLM behavior~\cite{coop, cocoop, tpt}. (2) Parameter-based methods either strategically update the pretrained parameters~\cite{miro,clipood,oh2024towards}, or learn new parameters by distillation~\cite{chen2024practicaldg,li2023distilling,addepalli2024leveraging}. (3) Feature-based methods either update the extracted features with additional modules~\cite{gao2024clip,unimos}, or build train-free cache modules with feature prototypes~\cite{zhang2022tip,tda,zhang2024dual}. From the view of transfer learning and heavily-studied transfer settings~\cite{zhuang2020comprehensive,weiss2016survey,tan2018survey}, we revisit these VLM-based methods and analyze the nuances derived from setting characteristics. Finally, we introduce mainstream benchmarks for different transfer settings and provide a rigorous performance comparison among VLM-based methods. As the strong abilities of VLMs bring significant boosts compared with single-modality methods, a benchmarking overview for VLM-based  transfer methods is urgently needed. 

\begin{figure*}[!t]
    \centering
    \includegraphics[width=0.93\textwidth]{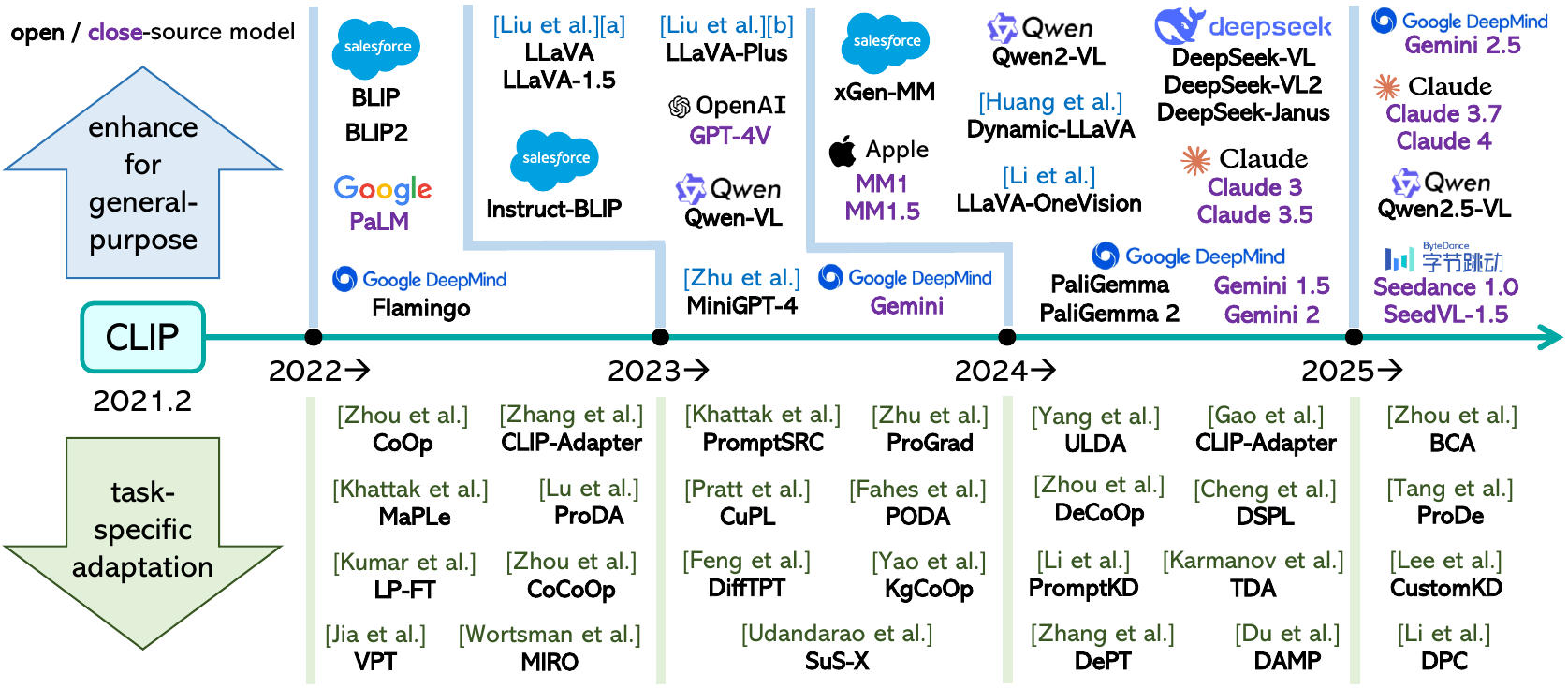}
    \vspace{-8pt}
    \caption{An overview of recent advances in vision-language systems, including adapting VLMs (lower part) and building general MLLMs (upper part).}
    \label{fig:overview}
    \vspace{-8pt}
\end{figure*}

The revolutionary breakthroughs in modern large language models (LLM)~\cite{bi2024deepseek,team2024qwen2,Qwen2023,achiam2023gpt} have further pushed the limits of VLMs. By connecting language-aligned vision encoders (e.g., CLIP's vision encoder) with LLMs and train with vast multimodal instruction-following data, the resultant vision-LLM, termed multimodal large language model (MLLM) in this survey, exhibits strong and vast generalization to a wider range of general visual-language tasks~\cite{lu2024deepseek,bai2023qwenvlversatilevisionlanguagemodel,liu2023visual,zhu2023minigpt}, including video understanding, visual question answering, image captioning, segmentation and recognition, etc. The development of general-purpose MLLMs is a fast-evolving field involving multiple top AI institutions, with new-generation products being released monthly. As MLLMs are another means of important vision-language model generalizable to various tasks, this survey includes the most typical and most up-to-date MLLMs to show practical and scaled applications of vision-language research. This survey summarizes general frameworks for building MLLMs, and discusses the model types, used training data, optimization objectives, and features of different MLLMs, aiming to provide readers with a comprehensive understanding of frontier advances in the field of vision-language research. An overview of recent advances in general vision-language research is presented in \cref{fig:overview}.
The contributions of this survey are summarized as follows:
\begin{enumerate}[leftmargin=26pt]
    \item This survey systematically reviews the research progress of knowledge transfer and generalize of VLMs, covering popular transfer settings including unsupervised domain adaptation, domain generalization, few-shot adaptation, test-time adaptation, etc. To the best of our knowledge, this paper is the first to investigate VLM-based methods from the perspective of model generalization.
    \item This survey identifies three key components within VLM architectures that are critical for knowledge transfer: prompt-based, parameter-based, and feature-based methods. A more fine-grained analysis is conducted to elucidate the specific techniques and adaptations employed within each transfer setting.
    \item This survey collects mainstream benchmarks for evaluating VLM-based generalization methods. A detailed performance comparison considering the generalization settings, model backbones and method design is provided, contributing a fair and comprehensive evaluation of current research advances.
    \item This survey includes VLMs enhanced and generalized by modern LLMs, termed MLLMs. A comprehensive summarization of the typical and cutting-edge MLLMs regarding their structures, used unimodal models, generalization to vast vision-language tasks, training data and objectives, is provided.
    \item This survey analyzes the key challenges that persist in current vision-language research and discusses potential directions for future exploration.
\end{enumerate}

The rest of this survey is organized as follows. \cref{sec:background} introduces the preliminaries of the studied VLMs, as well as the typical transfer learning settings involved in this survey. \cref{sec:prompt} introduces prompt-based methods divided by the prompting scenarios: Train-time prompting (\cref{sec:prompt_train}) and test-time prompting (\cref{sec:prompt_test}). \cref{sec:parameter} introduces parameter-based methods, including the robust fine-tuning of VLMs' parameters (\cref{sec:parameter_ft}) and distilling the pretrained knowledge to student model parameters (\cref{sec:parameter_kd}). \cref{sec:feature} introduces feature-level knowledge transfer, where the features are updated with learnable adapters (\cref{sec:feature_refine}) or compressed into train-free key-value caches (\cref{sec:feature_cache}). \cref{sec:experiment} describes benchmarks and evaluation results of the reviewed methods. \cref{sec:mllm} introduces how pretrained VLMs, e.g., CLIP, are generalized and enhanced with modern LLMs. \cref{sec:future} summarizes current research advances and discuss promising future directions in the vision-language field. 
\section{Background}
\label{sec:background}

\subsection{Preliminaries of Vision-Language Models}
\label{sec:background_pre}
Aligning with current research focus, this survey mainly discusses the \textit{two-tower VLM} as categorized in~\cite{zhang2024vision}, exemplified by CLIP~\cite{clip}. We start by introducing the pretraining objective, model structure and inference procedure of VLMs.

\begin{figure}[!t]
    \centering
    \includegraphics[width=0.5\textwidth]{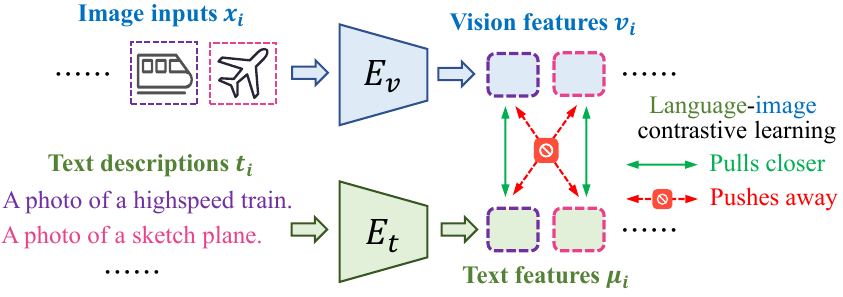}
    \vspace{-8pt}
    \caption{Illustration of the language-image contrastive learning procedure. $E_v$, $E_t$ are vision and language encoders, respectively.}
    \label{fig:contrastive}
    \vspace{-4pt}
\end{figure}

\textbf{Pretraining objective.} As shown in \cref{fig:contrastive}, VLMs are trained with abundant image-text pairs $\{(x_i, t_i)\}_{i=1}^{n}$, where $x_i$ are input images and $t_i$ are texts describing the image, e.g., \textit{A photo of a [object in the image]}~\cite{clip}. The VLM combines a vision and language encoder $E_v$ and $E_t$ to handle inputs from both modalities, outputting corresponding vision and text representations: $v_i = E_v(x_i)$, $\mu_i = E_t(t_i)$. The vision encoders are from the ResNet family~\cite{he2016deep} or the vision transformer family~\cite{dosovitskiy2020image}. Employing contrastive learning~\cite{zhang2022contrastive}, representations of matched vision and text pairs are pulled closer, while other pairs are pushed away. Given a batch of $B$ image-text pairs, the pretraining loss is defined as:
\begin{align}
    &\mathcal{L}_{\text{I2T}} = -\frac{1}{B} \sum_{i=1}^{B} \log \frac{\exp(v_i \cdot \mu_i) / \tau}{\sum_{j=1}^{B} \exp(v_i \cdot \mu_j) / \tau}, \\
    &\mathcal{L}_{\text{T2I}} = -\frac{1}{B} \sum_{i=1}^{B} \log \frac{\exp(\mu_i \cdot v_i) / \tau}{\sum_{j=1}^{B} \exp(\mu_i \cdot v_j) / \tau}, \\
    &\mathcal{L}_{\text{pre}} = \frac{1}{2} (\mathcal{L}_{\text{I2T}} + \mathcal{L}_{\text{T2I}}),
\label{eq:pretrain}
\end{align}
where $\tau$ is a learnable temperature parameter.

\textbf{Inference.} With vision and text encoders trained by \cref{eq:pretrain}, VLMs can make zero-shot inferences. Take the most general image classification as an example, the inference procedure is shown in \cref{fig:framework}. Given a set of $C$ candidate category names $\{\text{class}_c\}_{c=1}^C$, we can construct naive prompt inputs $t_c$ as in~\cite{clip}: $\text{A photo of a }\text{class}_c$. The text encoder then generates text representations $\{\mu_c\}_{c=1}^C$ corresponding to the class names. Given an image $x$ and its feature representation $v = E_v(x)$, the probability that image $x$ belongs to class $c$ is obtained by:
\begin{align}
    P_c(x) = \frac{\exp(\cos<v, \mu_c> / \tau)}{\sum_{i=1}^{C} \exp(\cos<v, \mu_i> / \tau)},
\label{eq:inference}
\end{align}
where $\cos<,>$ computes the cosine similarities. Note that based on the vision-language representation similarities, VLMs can be applied to various downstream tasks including object detection~\cite{zhong2022regionclip}, text retrieval~\cite{luo2022clip4clip}, etc. 

\textbf{Transferable modules.} As depicted in \cref{fig:framework}, this survey concludes the following three key VLM modules for knowledge transfer and generalization. 

\textit{(1) Prompts} are the text inputs $t_i$ for the language encoder $E_t$, describing the details of an image. However, the pretraining texts usually include more visual details~\cite{clip} while the naive prompts for downstream applications only include class names (\cref{eq:inference}), resulting in poor performance on more specialized tasks~\cite{coop}. Therefore, Zhou \textit{et al.}~\cite{coop} propose to learn the input prompts. As shown in \cref{fig:framework}, the prefix and suffix of prompts are learnable:
\begin{align}
    t_i = [U]_0, ..., [U]_k[\text{class}]_i[U]_{k+1}, ..., [U]_m,
\label{eq:prompt}
\end{align}
where $[U]_k$ are learnable prompt embeddings that replace the original texts. The prompts can either be appended to the start and end of a sentence. In addition to the prompts in \cref{eq:prompt}, there are various prompting paradigm and optimization objectives, which are elaborated in \cref{sec:prompt}.

\textit{(2) Parameters} in encoders $E_v, E_t$ are pretrained with \cref{eq:pretrain} and frozen for zero-shot predictions. However, with proper regularization the parameters can be updated for generalization to downstream tasks. A more practical scenario is to distill the VLM knowledge to smaller student models with parameters $\theta_\text{stu}$ for efficient deployments. Typical knowledge distillation optimizes student parameters by:
\begin{align}
    \theta_\text{stu} = \arg \underset{\theta_\text{stu}}{\min} \sum_{x} \mathcal{L}_{d}([P_0(x),...,P_C(x)], \texttt{stu}(x)),
\label{eq:distill}
\end{align}
where $\texttt{stu}(x)$ is the student output, and $\mathcal{L}_d$ is distance measurement, e.g., KL divergence.  \cref{sec:parameter} introduces more variants of fine-tuning and distillation methods. 

\textit{(3) Features} refer to the high-dimension vectors $\mu_i, v_i$ generated by the encoders. Inspired by the adapters for parameter-efficient fine-tuning~\cite{houlsby2019parameter}, the multimodal features can be efficiently refined with additional adapter modules. Take vision features as an example, the adapter $\Phi$ works by:
\begin{align}
    v_i^* = v_i + \alpha \cdot \Phi(v_i),
\label{eq:adapter}
\end{align}
where $v_i^*$ is the updated feature, $\alpha$ is a hyperparameter. As a more efficient alternative, key-value cache models can be built with the extracted features and corresponding model outputs, encompassing the model knowledge. 
Detailed discussions are in \cref{sec:feature}.

\subsection{Transfer Learning}
\label{sec:background_setting}
Transfer learning (TL) has been attracting research interests since more than a decade ago~\cite{torrey2010transfer}. TL enables deep models to share and reuse knowledge for similar tasks, enhancing annotation and computation efficiency in deep learning. This survey focuses on the following most frequently investigated TL settings in VLM-based methods.

\textbf{(1) Unsupervised domain adaptation}~\cite{cdan,li2018transfer,li2020maximum,li2021divergence} (UDA) aims to transfer knowledge from a labeled source domain to an unlabeled target domain with shifted distribution. Specifically, the source domain is denoted as $D_S=\{(x^S_i, y^S_i)\}_{i=1}^{n_S}$, and the target domain contains only unlabeled data $D_T=\{x^T_i\}_{i=1}^{n^T}$. The source and target domain follows different distributions: $P(X_S) \neq P(X_T)$.  The goal is to train a domain-invariant model $\varphi $ with $D_S$ and $D_T$ that minimizes prediction error $\mathcal{l}(\varphi(x^T), y^T)$ on the target domain. Typical single-modality methods either minimizes explicit measurements of domain gap~\cite{hu2015deep,yan2017mind} or implicitly aligns source and target features with adversarial training~\cite{dann, cdan}.

\textbf{(2) Domain generalization}~\cite{zhou2022domain,du2021learning} (DG) considers a more practical scenario where target data are \textit{unseen} during training, i.e., the model is only trained with labeled $D_S$. The model needs to learn source knowledge while retaining the ability to generalize to out-of-distributions (OOD). Single-modality methods aim to learn domain-invariant representations by metric learning~\cite{muandet2013domain,du2022energy}, adversarial training~\cite{li2018deep,zhao2020domain}, meta learning~\cite{li2018learning,balaji2018metareg}, and so on.

\textbf{(3) Test-time adaptation}~\cite{liang2025comprehensive} (TTA) adjusts model behavior at inference time with high efficiency. The source pretrained source model $\varphi$ is expected to produce accurate target predictions with minimum training. Since the target domain is unlabeled, typical methods adopt unsupervised measurements, e.g., entropy~\cite{wang2020tent} to assess or improve the prediction confidences. A similar setting is source-free domain adaptation~\cite{li2024comprehensive,li2022source} (SFDA). The difference is that SFDA assumes availability of the whole target domain for adaptation, while TTA only has access to data streams.

\textbf{(4) Few-shot learning}~\cite{wang2020generalizing,li2019leveraging} (FSL) allows limited annotation on target samples. Specifically, an \textit{N-way-K-shot} FSL problem includes $N$-category target training data with $K$ labeled samples for each class. FSL provides a data-efficient way for downstream adaptation, especially suitable for VLMs with vast base knowledge. 
The following method introductions are categorized according to the transferred modules and applied transfer settings introduced above. 
\section{Prompt-Based Methods}
\label{sec:prompt}
\begin{figure}[!t]
    \centering
    \includegraphics[width=0.47\textwidth]{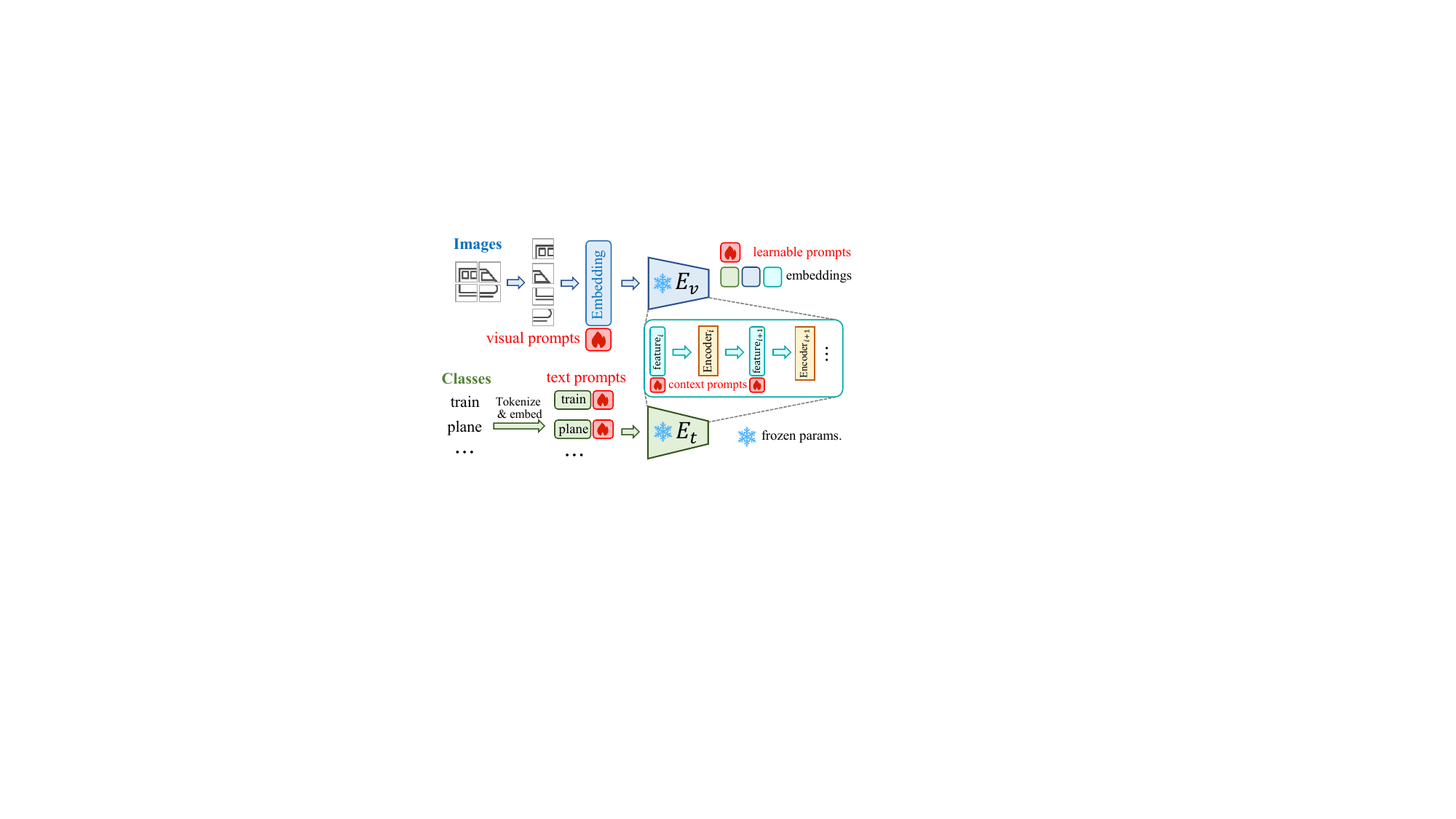}
    \vspace{-4pt}
    \caption{Illustrations of text, visual and context prompts.}
    \label{fig:prompt}
    \vspace{-8pt}
\end{figure}

As introduced in \cref{sec:background_pre}, prompts are in essence learnable parameters injected to the embeddings or features. In addition to the \textit{text prompts} appended to the word embeddings~\cite{coop}, researchers have proposed \textit{visual prompts}~\cite{vpt} and \textit{context prompts}~\cite{maple}. As depicted in \cref{fig:prompt}, the differences between these prompts are the applied locations. Text prompts complement the textual descriptions of the language branch. Visual prompts are appended to the image patches of vision transformer~\cite{dosovitskiy2020image}. Context prompts refine the intermediate features in both encoders. These prompting techniques exhibit distinct behaviors depending on the target tasks and optimization goals, which is elaborated as follows. 

\subsection{Prompting at Train Time}
\label{sec:prompt_train}
Proposed in \cite{coop}, train-time prompting has become a representative methodology for fine-tuning VLMs. An overview of prompt-based tuning methods covering the solved problem setting, prompt type and method description is presented in \cref{tab:prompt}. The major difference in the problem setting comes from the accessibility of labels, where FSL provides limited annotation for each class, while the rest settings are target-unlabeled.

\subsubsection{Few-Shot Prompting}
With labeled samples in FSL, prompt-tuning methods can adapt the VLMs to downstream tasks efficiently. Inspired by CoOp~\cite{coop}, there have been extensive explorations on the prompting style. While CoOp appends learnable prompts for text embeddings (type T), VPT~\cite{vpt} explores visual prompts to refine the tokenized input image patches for the vision transformer of CLIP (type V). Follow-up works~\cite{zang2022unified,xing2023dual,cho2023distribution} combine the strength of both modalities by simultaneously learning text and vision prompts. Rather than prompting \textit{before} encoders, MaPLe~\cite{maple} proposes to prompt intermediate features \textit{in} the encoders (type C). Different from the three types of prompts that applies to the representation space, Bahng \textit{et al.}~\cite{bahng2022exploring} propose to learn \textit{visible} prompts at the pixel level, an approach relevant to unadversarial training~\cite{salman2021unadversarial}. 

While the methods introduced above focus on the prompting position, there are attempts to improve the generalizability of the learned prompts. For better class-level base-to-new generalizability, CoCoOp~\cite{cocoop} enhances the class-agnostic prompts in CoOp by introducing a class-conditioned prompting meta-network, while methods~\cite{zhang2024dept,li2025dpc,zhou2024decoop} explicitly disentangles learnable prompts for base and new classes. Taking inspirations from robust fine-tuning~\cite{wortsman2022robust}, KgCoOp~\cite{yao2023visual} constrains learned prompts to be similar with default prompts to prevent overfitting. ProGrad~\cite{zhu2023prompt} prevents forgetting by aligning prompt update gradients. PromptSRC~\cite{khattak2023self} introduces prompt tuning regularizations through mutual agreement, self-ensembling and textual diversity. Identifying the limited informativeness of the default prompt in \cref{sec:background_pre}~\cite{pratt2023does}, ProText~\cite{khattak2025learning} introduces external knowledge from large language models (LLMs) like GPT3~\cite{brown2020language} to guide the prompt tuning process.

\subsubsection{Generalizable Prompting}
\label{sec:prompt_generalize}
Domain generalizable prompting aims to generalize learned prompts to unlabeled out-of-distributions (OOD), including the setting of DG and UDA. One line of research aims to learn \textit{domain-invariant} prompts that is generalizable across different domains, as well as \textit{domain-specific} prompts that encode domain knowledge~\cite{daprompt,singha2023ad,bose2024stylip,damp,phan2024enhancing,bose2024stylip}. By modifying \cref{eq:prompt}, the domain prompts are presented as:
\begin{align}
    t_c^d = [U]_0, ..., [U]_k[V]_{k+1}^d, ..., [V]^d_m[\text{class}]_c,
\label{eq:domain_prompt}
\end{align}
where $[U]_i$ are domain-invariant prompts, $[V]_i^d$ are specialized prompts for domain $d$, $d \in \{s,t\}$ represent source or target domain, and $c$ is the index of category. The prompts in \cref{eq:domain_prompt} are trained with labeled source data, and pseudo-labeled target data in UDA~\cite{daprompt, damp}. The pseudo labels are generally obtained via VLMs' high-quality zero-shot predictions, which contributes to the significant performance boosts, as detailed in \cref{sec:experiment}. STYLIP~\cite{bose2024stylip} trains additional modules to model domain style. DAMP~\cite{damp} fosters knowledge and alignments between different modality encoders.

\begin{table*}[!t]
    \caption{Prompt-based train-time generalization methods for VLMs. T, V, C represent text, visual, context prompt types, respectively.}
    \centering
    \label{tab:prompt}
    \vspace{-8pt}
    \resizebox{\linewidth}{!}{
    \setlength{\tabcolsep}{5pt}
    \begin{tabular}{@{}p{2.1cm}cccl@{}}
    \toprule
    \textbf{Method} & \textbf{Venue} & \textbf{Setting} & \textbf{Type} & \textbf{Brief description} \\ \midrule
    LDFS~\cite{yan2024enhancing} & Arxiv & DG & T & A   plug-and-play feature synthesis method to synthesize new domain features. \\
    DSPL~\cite{cheng2024disentangled} & CVPR'24 & DG & V,T & Disentangles text prompts using LLM then learn domain-specific and invariant prompts. \\
    ODG-CLIP~\cite{singha2024unknown} & CVPR'24 & DG & T & Generates OOD samples with diffusion models for training prompts. \\
    Xiao \textit{et al.}~\cite{xiao2024any} & CVPR'24 & DG & V,T & A probabilistic inference framework that considers both training and test distributions. \\
    Hao \textit{et al.}~\cite{hao2024quantized} & ECCV'24 & DG & C & Utilizes quantization error as a kind of noise to explore   quantization for regularization. \\
    SPG~\cite{bai2024soft} & ECCV'24 & DG & T & Trains instance-specific soft prompts for unseen target domains. \\
    CoOPood~\cite{zhang2024amend} & ICML'24 & DG & T & A   fine-grained prompt tuning method that mitigates spurious correlation. \\
    OGEN~\cite{zang2024overcoming} & ICLR'24 & DG & T & Synthesizes OOD features to regularize decision boundaries between ID and OOD data. \\
    STYLIP~\cite{bose2024stylip} & WACV'24 & DG & T & Learns   domain-agnostic prompts with a set of style projectors. \\
    MetaPrompt~\cite{bosemeta} & TMLR'25 & DG & T & Learns   prompts to detect unknown class  using   meta-learning with momentum updates. \\
    DAPrompt~\cite{daprompt} & TNNLS'23 & UDA & T & Learns  domain information with domain-specific and domain-agnostic prompts. \\
    AD-CLIP~\cite{singha2023ad} & ICCVW'23 & UDA & T & Learns   domain-invariant prompts by conditioning on image style and content features. \\
    DAMP~\cite{damp} & CVPR'24 & UDA & C,T & Learns   domain-invariant semantics by mutually aligning visual and textual   embeddings. \\
    PDA~\cite{bai2024prompt} & AAAI'24 & UDA & V,T & Learns  a base and alignment branch to integrate class-related and cross-domain features. \\
    Shi \textit{et al.}~\cite{shi2024clip} & IJCNN'24 & UDA & T & Trains   CLIP's prompts and an image encoder with data augmentation. \\
    FUZZLE~\cite{shi2024unsupervised} & TFS'24 & UDA & T & Integrates  fuzzy C-means clustering and a fuzzy vector during prompt learning. \\
    MPA~\cite{chen2023multi} & NeurIPS'23 & MSDA & T & Minimizes the domain gap between each source-target domain pair by prompt learning. \\
    PGA~\cite{phan2024enhancing} & NeurIPS'24 & MSDA & T & Trains multi-domain prompts by solving a multi-objective optimization problem. \\
    UPT~\cite{zang2022unified} & Arxiv & FSL & V,T & Learns   an additional network to simultaneously optimize vision and text prompts. \\
    CoOp~\cite{coop} & IJCV'22 & FSL & T & Learns a class-agnostic text prompt to adapt VLMs to downstream tasks. \\
    CoCoOp~\cite{cocoop} & CVPR'22 & FSL & T & Proposes   to learn class-conditioned prompts with a meta-network. \\
    ProDA~\cite{lu2022prompt} & CVPR'22 & FSL & T & Learns  from diverse prompts to handle the varying visual representations. \\
    VPT~\cite{vpt} & ECCV'22 & FSL & V & Learns  visual prompts for the input of  vision transformer. \\
    MaPLe~\cite{maple} & CVPR'23 & FSL & C & Learns context prompts in both vision and text encoder. \\
    PromptSRC~\cite{khattak2023self} & CVPR'23 & FSL & V,T & Promotes consistency between prompted and pretrained features to prevent forgetting. \\
    KgCoOp~\cite{yao2023visual} & CVPR'23 & FSL & T & Regulates   the learned prompts to be similar with the hand-crafted prompts. \\
    ProGrad~\cite{zhu2023prompt} & ICCV'23 & FSL & T & Only  updates prompts whose gradient align with the general knowledge. \\
    DAPT~\cite{cho2023distribution} & ICCV'23 & FSL & V,T & Learns multimodal prompts by finding the appropriate distribution in each   modality. \\
    DPT~\cite{xing2023dual} & TMM'23 & FSL & V,T & Simultaneously learns the
    visual and text prompts from the ends of both encoders. \\
    DePT~\cite{zhang2024dept} & CVPR'24 & FSL & T & Decouples and preserves  base-specific knowledge in prompts. \\
    DeCoOp~\cite{zhou2024decoop} & ICML'24 & FSL & T & Tunes prompts on base classes and evaluates on a combination   of base and new classes. \\ 
    ProText~\cite{khattak2025learning} & AAAI'25 & FSL & C,T & Learns prompts from the rich contextual   knowledge in LLM data. \\
    ZIP~\cite{park2025zip} & ICLR'25 & FSL & C & Reduces the problem dimensionality and the variance of zeroth-order gradient estimates. \\ 
    DPC~\cite{li2025dpc} & CVPR'25 & FSL & T & Decouples   the learning of base and new tasks at prompt level. \\\bottomrule
    \end{tabular}}
    \vspace{-5pt}
\end{table*}

As a typical category of DG methods~\cite{wang2022generalizing}, data generation techniques synthesize diverse labeled data from unseen domain to enhance the model's robustness~\cite{qiao2020learning,zhou2020learning,li2021simple}. For VLM-based methods, data generation can be achieved at the feature level with learnable augmenters in a meta-learning style~\cite{yan2024enhancing,zang2024overcoming,bosemeta}, or using more advanced generative techniques like stable diffusion~\cite{singha2024unknown}. As a special kind of augmentation, Hao \textit{et al.}~\cite{hao2024quantized} observe that random noise can contribute to model robustness, and propose to utilize \textit{quantization error} for regularization. 

As a variant of UDA, multi-source UDA (MSDA)~\cite{peng2019moment,li2024agile} introduces multiple labeled source domains with different distribution, and the model is expected to aggregate all the domain knowledge to enhance target performance. Intuitively, MSDA can be solved by training prompts in \cref{eq:domain_prompt} for each domain, i.e., $d \in \{s_0,s_1,...,s_N,t\}$~\cite{zhu2023prompt,phan2024enhancing}. With the pretrained general knowledge, VLM-based UDA methods can be extended for MSDA by viewing all source domains as a whole domain with mixed distribution~\cite{damp}.

\subsection{Prompting at Test Time}
\label{sec:prompt_test}
Prompting for TTA requires on-the-fly adjustments of the VLM in an unsupervised manner. Current prompt-based TTA methods mostly base on TPT~\cite{tpt}. Denote the frozen VLM as $\varphi(x;\mathbf{t})$ that takes image input $x$ and prompts $\mathbf{t} = \{t_i\}_{i=1}^C$ of all $C$ classes. The TPT loss is defined as: 
\begin{align}
    \mathcal{L}_{\text{tpt}} = \text{Ent}(\frac{1}{B}\sum_{i=1}^B \varphi(x_i;\mathbf{t})),
\label{eq:tpt}
\end{align}
where $\{x_i\}_{i=1}^B$ are augmented views with high confidence given every original input $x$, $\text{Ent}(\mathbf{y})=-\sum_{c=1}^{C}y_c \log y_c$ is Shannon entropy~\cite{shannon1948mathematical}. Intuitively, \cref{eq:tpt} maximizes the consistency among all reliable augmented views of a test input. Follow-up methods seek to improve this paradigm with further regularization or advanced augmentation techniques. \cref{tab:prompt_tta} summarizes these methods and provides illustrative training losses. Note the losses are refined for better readability. Refer to the original papers for details.


\begin{table}[t]
    \caption{Test-time prompt tuning methods for VLMs.}
    \centering
    \label{tab:prompt_tta}
    \vspace{-5pt}
    \resizebox{\linewidth}{!}{
    \setlength{\tabcolsep}{5pt}
    \begin{tabular}{@{}ll@{}}
    \toprule
    \textbf{Method} & \textbf{Training loss} \\ \midrule
    TPT~\cite{tpt} & $\mathcal{L}_{\text{tpt}}$: Test-time prompting loss in \cref{eq:tpt}. \\ \midrule
    PromptAlign~\cite{abdul2023align}  & $\mathcal{L}_{\text{tpt}}$+$\| \varepsilon(\varphi(x;t)), \hat{\varepsilon} \|_1 + \| \sigma(\varphi(x;t)), \hat{\sigma} \|_1 $. \\\midrule
    DiffTPT~\cite{difftpt} & $\mathcal{L}_{\text{tpt}}$ with augmentations from diffusion model. \\\midrule
    \begin{tabular}[c]{@{}l@{}} Swap- \\Prompt~\cite{ma2023swapprompt} \end{tabular}  & \begin{tabular}[c]{@{}l@{}} $\mathcal{L} = \mathcal{l}_{\text{ce}}(\varphi(x;t'), \hat{y}) $ + $\mathcal{l}_{\text{ce}}(\varphi(x';t'), \hat{y})$ + \\ $\mathcal{l}_{\text{ce}}(\varphi(x;t'), \varphi(x';t)) $ + $\mathcal{l}_{\text{ce}}(\varphi(x';t'), \varphi(x;t))$. \end{tabular} \\\midrule
    C-TPT~\cite{yoon2024c} & $\mathcal{L}_{\text{tpt}}$+ $\frac{1}{C} \sum_{c=1}^{C} \| \mathbf{t}_{\text{centroid}} - t_c \|_2$. \\\midrule
    DynaPrompt~\cite{xiao2025dynaprompt}  & $\mathcal{L}_{\text{tpt}}$ on instance-relevant prompts. \\\midrule
    R-TPT~\cite{sheng2025r} & $\mathcal{L}= \frac{1}{B} \sum_{i=1}^{B} \text{Ent}(\varphi(x_i;t))$. \\\midrule
    O-TPT~\cite{sharifdeen2025tpt} & $\mathcal{L}_{\text{tpt}}$+$\| \mathbf{E}\mathbf{E}^\top - \mathbf{I} \|_2^2$, $\mathbf{I}$ is the identity matrix. \\ \bottomrule
    \end{tabular}}
    \vspace{-8pt}
\end{table}

The regularization-based TTA methods in \cref{tab:prompt_tta} include: PromptAlign~\cite{abdul2023align} additionally aligns the means $\varepsilon$ and variances $\sigma$ of the learned prompts with the source dataset statistics $\hat{\varepsilon}, \hat{\sigma}$ using $L_1$ norm. C-TPT~\cite{yoon2024c} learns prompts similar to the text feature centroid $\mathbf{t}_{\text{centroid}}=\frac{1}{C}\sum_{c=1}^{C} t_c$. R-TPT~\cite{sheng2025r} modifies the TPT loss in \cref{eq:tpt} by only minimizing point-wise entropy. O-TPT~\cite{sharifdeen2025tpt} applies orthogonal regularization on text feature matrix $\mathbf{E}$, where the $i_{\text{th}}$ row $E_{i}$ represents prompt embedding of the $i_{\text{th}}$ class. SwapPrompt~\cite{ma2023swapprompt} introduces target prompts $t$ and online prompts $t'$ that updates the target prompts in an Exponential Moving Average (EMA) manner, and encourages prediction consistencies among different prompts and augmented inputs $x'$. Other methods include DiffTPT~\cite{difftpt} that replaces the standard augmentations, e.g., random cropping and flipping in~\cite{tpt}, with images generated by diffusion model. DynaPrompt~\cite{xiao2025dynaprompt} only selects and optimizes relevant prompt for each test sample.

\subsection{Discussion}
Prompt tuning has become the mainstream approach to generalize VLMs in various cases. Categorized by prompting positions, prompts can be learned on text, visual embeddings and intermediate features. One can also tailor the prompts into domain-invariant and domain-specific parts to solve cross-domain generalization tasks. The computation efficiency of prompt tuning makes it suitable for test-time adaptation with proper calibration. Apart from generalization tasks, prompting can be applied as a task-agnostic tuning technique for VLMs~\cite{luddecke2022image,vidit2023clip,jiang2023clip}. Despite their effectiveness, prompt tuning may fall behind full fine-tuning in certain scenarios like large distribution gaps~\cite{lai2023padclip}. 
\section{Parameter-Based Methods}
\label{sec:parameter}
Prompt-based methods introduced in \cref{sec:prompt} learn additional prompts  with original parameters frozen, which may not be sufficient for tasks with significant domain gap~\cite{lai2023padclip}. Fine-tuning on pretrained models is an effective approach for transfer learning~\cite{tan2018survey,zhang2024vision}. However, the delicate pretrained multimodal representations in VLMs require more strategical fine-tuning processes. Inspired by knowledge distillation (KD)~\cite{hinton2015distilling}, a more practical transfer approach distills the knowledge in VLMs to a smaller student for efficient deployment or data-sensitive cases. The two lines of study either modify the original parameters, or introduce new parameters, extending the ability of VLMs. \cref{fig:parameter} illustrates typical fine-tuning and distillation procedures.

\subsection{Fine-Tuning Methods}
\label{sec:parameter_ft}
\begin{figure}[!t]
    \centering
    \includegraphics[width=0.47\textwidth]{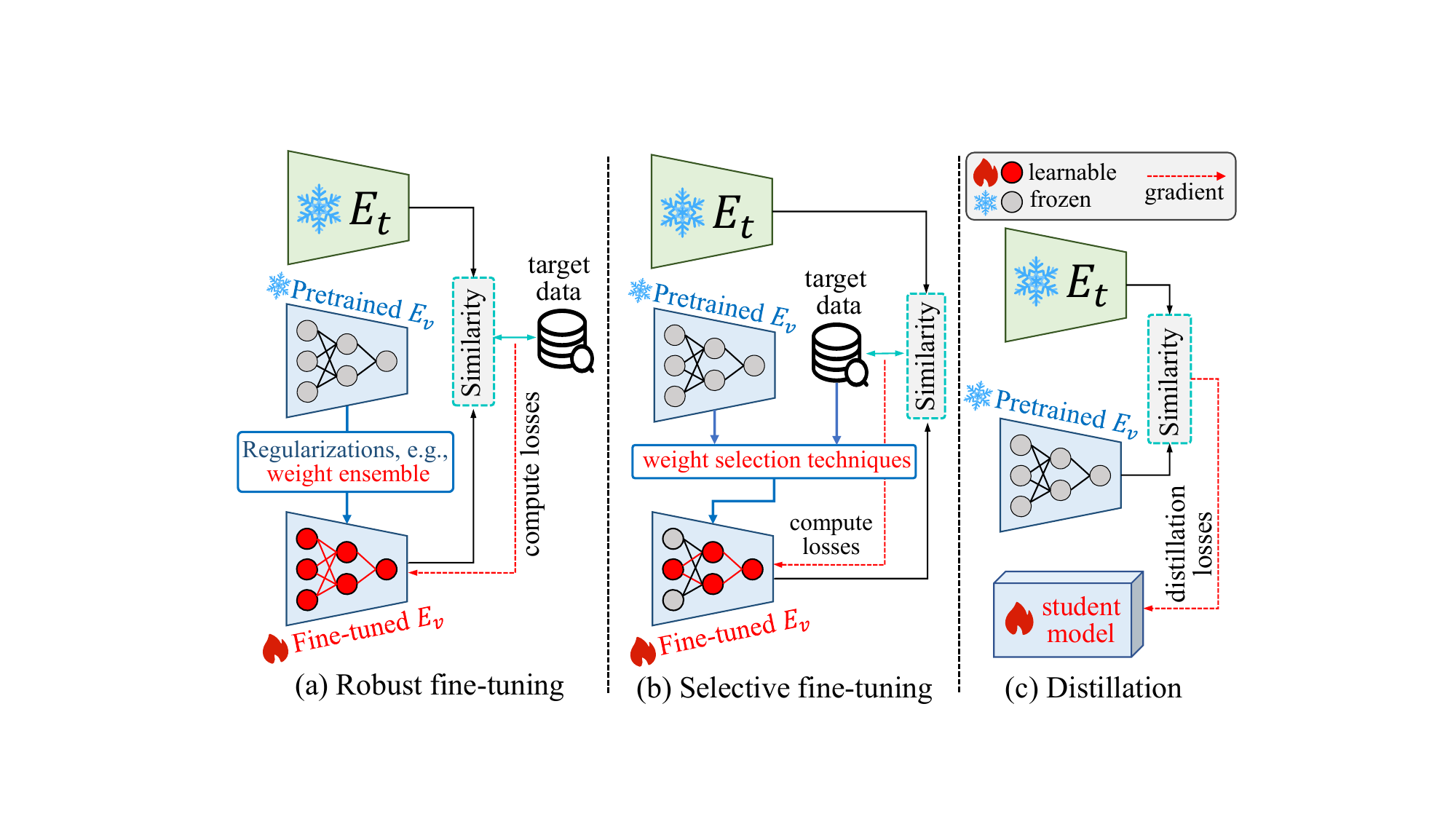}
    \vspace{-4pt}
    \caption{Illustrations of parameter-based methods. For clarity, only the vision encoder is transferred. (a) Robust fine-tuning methods adopt regularizations like weight ensemble to prevent overfitting. (b) Selective fine-tuning methods choose task-relevant weights to update. (c) Distillation methods learn a new student model from pretrained VLMs' outputs.}
    \label{fig:parameter}
    \vspace{-6pt}
\end{figure}

\cref{tab:parameter} includes representative fine-tuning-based knowledge transfer methods for VLMs, mainly covering the setting of DG and TTA. Depending on the tuning process, these methods can be categorized as robust fine-tuning (FT-R) and selective fine-tuning (FT-S), as illustrated in \cref{fig:parameter}(a),(b).

\subsubsection{Robust Fine-Tuning}
\label{sec:parameter_robust_ft}
Traditional fine-tuning focuses more on acquiring new target knowledge, while the tuning process of VLMs highlights the challenge to \textit{preserve} pretrained knowledge. Extensive researches have revealed that naive fine-tuning leads to severe forgetting and damages of pretrained knowledge~\cite{kumar2022fine,gao2024clip,zhang2022tip,lai2023padclip}, which inspire the explorations on \textit{robust} fine-tuning, aiming to adapt to target tasks without compromising the models' original abilities.

As categorized in \cref{tab:parameter}, robust fine-tuning (FT-R) methods are mainly designed for DG tasks for better generalization ability. First revealed in \cite{keskar2016large}, finding \textit{flat minima} in losses has become a mainstream DG framework. Izmailov\textit{et al.}~\cite{izmailov2018averaging} proposes a simple baseline stochastic weight averaging (SWA) that finds flat minima by averaging trained model weight snapshots every K epochs. Cha \textit{et al.}~\cite{swad} improves SWA by introducing iteration-granularity ensemble when the validation loss is low, namely Stochastic Weight Averaging Densely (SWAD). SWA and SWAD has been successfully applied to VLMs like CLIP and inspire further researches on weight-ensemble-based robust fine-tuning of VLMs. WiSE-FT~\cite{wortsman2022robust} proposes to combine specifically the pretrained weights and fine-tuned weights. Instead of preserving zero-shot weights via explicit mixing, follow-up works~\cite{miro,mao2024context} implicitly enforces trained weights to be similar with zero-shot ones.  CLIPood~\cite{clipood} performs adaptive model ensemble with ensemble weights sampled from Beta distribution. DART~\cite{jain2023dart} assembles model weights trained via different augmentations of input images, while WATT~\cite{watt} obtain diverse model weights for ensemble via different prompt templates.

In addition to weight-level generalization, more diverse strategies have been explored to maintain zero-shot capabilities during fine-tuning. LP-FT~\cite{kumar2022fine} proposes to fine-tune with prediction heads initialized with frozen zero-shot weights. FLYP~\cite{goyal2023finetune} preserves the pretraining \textit{strategy} of zero-shot weights for fine-tuning. PADCLIP~\cite{lai2023padclip} prevents forgetting of zero-shot knowledge by adopting low and finely controlled learning rates. LipsumFT~\cite{nam2024lipsum} connects zero-shot and fine-tuned vision encoder weights by the frozen text encoder. CaRot~\cite{oh2024towards} adopts a new multimodal contrastive loss to promote larger smallest singular values then performs weight ensemble. UEO~\cite{ueo} extends the information maximization loss in \cite{liang2020we} by minimizing in-distribution samples' information entropy while maximizing the OOD samples' entropy. 

\begin{table*}[!t]
    \caption{Parameter-based transfer methods for VLMs. FT-S, -R represent selective and robust fine-tuning methods. KD indicates knowledge distillation.}
    \centering
    \label{tab:parameter}
    \vspace{-8pt}
    \resizebox{\linewidth}{!}{
    \setlength{\tabcolsep}{4.9pt}
    \begin{tabular}{@{}p{2.03cm}cccl@{}}
    \toprule
    \textbf{Method} & \textbf{Venue} & \textbf{Setting} & \textbf{Type} & \textbf{Brief description} \\ \midrule
    SWAD~\cite{swad} & NeurIPS'21 & DG & FT-R & Finds   flatter minima with a dense and overfit-aware stochastic weight sampling strategy. \\
    WiSE-FT~\cite{wortsman2022robust} & CVPR'22 & DG & FT-R & Assembles   the weights of pre-trained and fine-tuned models. \\
    MIRO~\cite{miro} & ECCV'22 & DG & FT-R & Maximizes   the mutual information between fine-tuned and pre-trained features. \\
    LP-FT~\cite{kumar2022fine} & ICLR'22 & DG & FT-R & Proposes   fine-tuning with classification head initialized by linear-probing. \\
    CLIPood~\cite{clipood} & ICML'23 & DG & FT-R & Assembles   model weights by Beta distribution and adopts a margin softmax loss   for training. \\
    FLYP~\cite{goyal2023finetune} & CVPR'23 & DG & FT-R & Proposes to use the exact pretraining loss for downstream fine-tuning. \\
    DART~\cite{jain2023dart} & CVPR'23 & DG & FT-R & Combines   the weights of models trained with differently augmented inputs. \\
    UEO~\cite{ueo} & ICML'24 & DG & FT-R & Leverages instance-level confidence to optimize through entropy. \\
    SAFT~\cite{nguyen2024saft} & ECCV'24 & DG & FT-S & Selectively   updates a small subset of parameters whose gradient magnitude is large. \\
    CaRot~\cite{oh2024towards} & NeurIPS'24 & DG & FT-R & Proposes   to fine-tune VLMs by enforcing a larger smallest singular value. \\
    LipsumFT~\cite{nam2024lipsum} & ICLR'24 & DG & FT-R & Regularizes   the Energy Gap from language model outputs for random texts during   fine-tuning. \\
    CAR-FT~\cite{mao2024context} & IJCV'24 & DG & FT-R & Minimizes   KL divergence between zero-shot and learned prompt weights during   fine-tuning. \\
    CLIP-TD~\cite{wang2022clip} & Arxiv & DG & KD & Distills   knowledge from CLIP into existing architectures using a dynamically weighted   objective. \\
    APD~\cite{luo2024adversarial} & Arxiv & DG & KD & Multimodal   knowledge KD framework that trains student model with adversarial   examples. \\
    DFKD~\cite{xuan2023distilling} & MM'23 & DG & KD & Distills   a student model for distribution-agnostic downstream tasks with given   categories. \\
    Li \textit{et al.}~\cite{li2023distilling} & ICCV'23 & DG & KD & Imitates   teacher's visual representations to promote coherence in   vision-language alignment.  \\
    RISE~\cite{huang2023sentence} & ICCV'23 & DG & KD & Regularizes the student's learned   representations to be close to the teacher's. \\
    DAD~\cite{zhou2023distilling} & NeurIPS'23 & DG & KD & Leverages   VLM teacher to generate adversarial examples and a VQGAN to discretize them. \\
    SCI-PD~\cite{chen2024practicaldg} & CVPR'24 & DG & KD & Transfers   knowledge from VLMs to lightweight vision models with improved  robustness. \\
    VL2V-ADiP~\cite{addepalli2024leveraging} & CVPR'24 & DG & KD & Aligns   vision and text modalities between the teacher and student model then   perform KD. \\
    PromptKD~\cite{li2024promptkd} & CVPR'24 & DG & KD & Domain-specific prompt-based knowledge distillation with unlabeled target data. \\
    KDPL~\cite{mistretta2024improving} & ECCV'24 & DG & KD & A KD framework that is applicable to prompt tuning methods of VLMs. \\
    PADCLIP~\cite{lai2023padclip} & ICCV'23 & UDA & FT-R & Fine-tunes  CLIP with adjusted learning rates to prevent forgetting, and debias pseudo   labels. \\
    CLIP-Div~\cite{zhu2024clip} & Arxiv & UDA & KD & Learns a feature extractor and classifier with language-guided pseudo labels. \\
    CMKD~\cite{zhou2024unsupervised} & TCSVT'24 & UDA & KD & Leverages   VLMs as teacher models to guide the learning process in the target domain. \\
    CustomKD~\cite{lee2025customkd} & CVPR'25 & UDA & KD & Customizes   the teacher features to reduce discrepancy between teacher   and student models. \\ 
    RLCF~\cite{zhao2024test} & ICLR'24 & TTA & KD & Adopts   reinforcement learning to improve TTA with CLIP feedback to rectify the model   outputs. \\
    CLIP-OT~\cite{mishra2024words} & Arxiv & TTA & FT-S & Adopts  optimal transfer to integrate and distill multiple template knowledge. \\
    SAR~\cite{niu2023towards} & ICLR'23 & TTA & FT-S & Removes  samples with large gradients and encourages model weights towards flat minimum. \\
    WATT~\cite{watt} & NeurIPS'24 & TTA & FT-S,R & Assembles weights learned from different text prompt templates. \\
    DPLOT~\cite{yu2024domain} & CVPR'24 & TTA & FT-S & Selects  and trains   domain-specific model blocks with paired-view pseudo labeling. \\
    CLIPArTT~\cite{hakim2025clipartt} & WACV'25 & TTA & FT-S & Adapts norm layers by combining multiple class prompts as pseudo labels. \\
    TTL~\cite{imam2025test} & WACV'25 & TTA & FT-S & A PEFT TTA method that updates the attention weights of the transformer encoder. \\
    POUF~\cite{tanwisuth2023pouf} & ICML'23 & SFDA & FT & Directly fine-tunes the model or prompts on the unlabeled target data. \\
    RCL~\cite{chen2024empowering} & Arxiv & SFDA & KD & Distillation from   multiple VLMs with reliability-driven learning. \\
    ViLAaD~\cite{tarashima2025vilaad} & Arxiv & SFDA & KD & A   framework that integrates VLMs as a powerful initialization for target   adaptation. \\
    DALL-V~\cite{zara2023unreasonable} & ICCV'23 & SFDA & KD & Distills  the rich multimodal knowledge in VLMs to a student model tailored for the   target. \\
    DIFO~\cite{tang2024source} & CVPR'24 & SFDA & KD & First customizes a VLM from the target model then distills its knowledge to the target model. \\
    Zhan \textit{et al.}~\cite{zhan2024towards} & IJCAI'24 & SFDA & KD & Fosters collaboration between the source trained model   and the VLM. \\
    Colearn++~\cite{zhang2025source} & IJCV'25 & SFDA & KD & A co-learning strategy that generates  reliable   pseudo labels by pretrained models for finetuning. \\
    ProDe~\cite{tang2024proxy} & ICLR'25 & SFDA & KD & A proxy denoising mechanism that corrects VLM's predictions for domain-invariant KD. \\\bottomrule
\end{tabular}}
\end{table*}

\subsubsection{Selective Fine-Tuning}
\label{sec:parameter_partial_ft}
While robust full fine-tuning methods introduced in \cref{sec:parameter_robust_ft} are effective for generalization, they are usually computationally intensive thus cannot satisfy the fast adaptation and inference needs in TTA. Therefore, selective fine-tuning (FT-S) is proposed to adapt VLMs in limited response time. These methods either adjust certain statistics in model to align with the distribution of input data, or find the most task-relevant parameters to fine-tune. 

Batch normalization (BN) layers~\cite{ioffe2015batch} play an important part in ensuring fast convergence and reducing covariate shifts. Prior works~\cite{li2016revisiting,schneider2020improving,you2021test} have discovered the effectiveness of simply adapting the statistics in BN layers for model generalization. Such finding further benefits the test-time adjustments of VLMs by replacing full parameter tuning with statistics tuning~\cite{mishra2024words,watt,hakim2025clipartt}. Rather than BN layers, Imam \textit{et al.}~\cite{imam2025test} choose to update attention layers following current parameter-efficient fine-tuning advances~\cite{hu2022lora}, and Niu \textit{et al.}~\cite{niu2023towards} select batch-agnostic norm layers like group norm and layer norm.  

The above introduced methods fine-tune a fix subset of model parameters, neglecting task specifics. The principles of selective fine-tuning~\cite{ding2023parameter,yang2022deep} are then utilized for VLMs, where model weights contributing more to the target task are updated. Furthermore, recent studies~\cite{aghajanyan2020intrinsic,arora2018stronger} have shown that reduced fine-tuned parameters contribute to better generalization abilities, confirming the effectiveness of selective fine-tuning. DPLOT~\cite{yu2024domain} explores such parameters by dividing the whole model into blocks then performs selection. SAFT~\cite{nguyen2024saft} selects important parameters by evaluating the gradients of cross-entropy losses with respect to the parameters. Rather than selecting parameters, SAR~\cite{niu2023towards} selects more reliable test samples for adaptation.

\subsection{Generalizable Distillation}
\label{sec:parameter_kd}
Fine-tuning methods in \cref{sec:parameter_ft} need to access the model parameters for computing gradient, which is not viable for black-box settings~\cite{ouali2023black,liang2022dine} where the pretrained models are only accessed through API calls. Furthermore, the increasing scale of pretrained foundation models hinder the deployments on edge devices, which motivates distilling smaller student models with similar abilities with the large teacher model~\cite{addepalli2024leveraging}. As summarized in \cref{tab:parameter}, knowledge distillation (KD) methods aim to purify domain generalizable student models, or utilize the teacher knowledge in VLMs to refine pretrained source models.

\textbf{Domain generalizable distillation.}
In the context of vision language distillation depicted by \cref{eq:distill}, the teacher model is the pretrained VLMs and the student model is generally a smaller vision model. The goal is to distill the knowledge in VLMs to the student model while also enhancing its OOD abilities. Similar to robust fine-tuning, generalizable distillation follows the principle of \textit{preserving pretrained knowledge} of VLMs~\cite{li2023distilling,huang2023sentence,addepalli2024leveraging}. To achieve this, VL2V-ADiP~\cite{addepalli2024leveraging} first aligns the representation space between the student model and the VLM's image encoder before distillation. RISE~\cite{huang2023sentence} constrains the distilled representations to be close to the pretrained ones. CustomKD~\cite{lee2025customkd} aligns features to overcome the model discrepancies brought by different structures, scales, etc. PromptKD~\cite{li2024promptkd} distills knowledge to a smaller VLM by preserving the teacher's pretrained text features. Adversarial distillation methods~\cite{luo2024adversarial,zhou2023distilling} introduce adversarial examples to enhance the robustness of student models. In addition to typical DG, SCI-PD~\cite{chen2024practicaldg} considers a more challenging open-set DG setting where target data might include novel classes. There are also researches on distilling for UDA~\cite{zhu2024clip,zhou2024unsupervised,lee2025customkd}. RLCF~\cite{zhao2024test} proposes to adopt pretrained VLMs as reward models for test-time adaptation with reinforcement learning.

\textbf{Distillation-guided source model refinement.}
The distillation methods for DG introduced above generally distill knowledge into an independently initialized task-irrelevant student model. The source-free domain adaptation (SFDA) setting~\cite{liang2025comprehensive} provides a source-trained student model and unlabeled target data. The goal is to adapt the student model to accurately predict target data. Single-modality methods refine the source model with unsupervised measurements like pseudo labels~\cite{liang2020we,dong2021confident}, but cannot achieve satisfactory results due to their low accuracies. VLMs provide a novel approach by distilling the pretrained knowledge on target data to the source model to solve SFDA, as shown in \cref{tab:parameter}. Pseudo-label-based methods~\cite{tang2024proxy,tarashima2025vilaad,zhang2025source} aim to generate more reliable pseudo labels for target data utilizing clustering~\cite{tarashima2025vilaad,zhang2025source} or proxy denoising~\cite{tang2024proxy}. RCL~\cite{chen2024empowering} exploits multiple LLMs for more accurate pseudo labels and then transfer knowledge in a curriculum learning framework. Other methods explicitly distill the teacher predictions from target-customized VLMs~\cite{tang2024source} via adapters~\cite{zara2023unreasonable} or prompts~\cite{zhan2024towards}. Yu \textit{et al.}~\cite{yu2024select} further explore the distillation methods that prevent forgetting in continual learning~\cite{hou2019learning,riemer2019learning}.

While the methods introduced above mainly focus on \textit{distribution-level} generalization for image classification tasks, KD is also applicable for \textit{task-level} knowledge transfer, e.g., distilling VLMs for object detection~\cite{zou2023object} or semantic segmentation~\cite{guo2018review} tasks. The goal is to exploit and transfer the vast general knowledge in VLMs for downstream data-scarce or open-vocabulary scenarios~\cite{gu2021open,ma2022open,bangalath2022bridging} by bridging the granularity gap between image-level pretraining of VLMs and pixel-level target tasks~\cite{bangalath2022bridging,xu2022simple,ghiasi2022scaling,zhou2022extract}.

\subsection{Discussion}
Compared with prompt-based methods, parameter-based methods provide more flexibility for generalizing VLMs according to various problem settings, computation constraints, data availability, etc. Introduced in \cref{sec:parameter_ft}, fine-tuning methods extend the traditional pretraining-finetuning framework to VLMs by balancing the learning of target knowledge and preservation of zero-shot knowledge. Two major research lines include robust fine-tuning that regulates the updated parameters, and selective fine-tuning that finds crucial parameters to learn. Distillation methods in \cref{sec:parameter_kd} introduce more flexible transfer methods that suit various practical scenarios and tasks.
\begin{figure}[!t]
    \centering
    \includegraphics[width=0.46\textwidth]{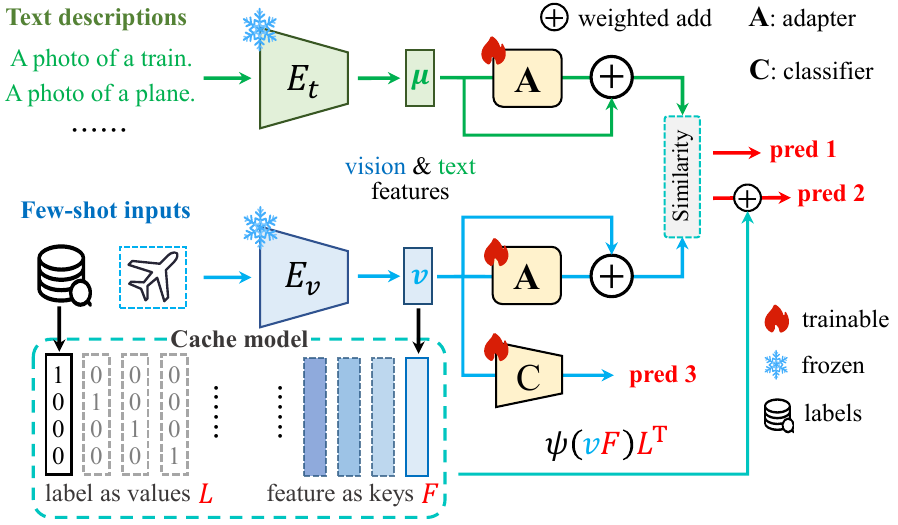}
    \vspace{-8pt}
    \caption{Illustrations of feature-based methods, where \textbf{pred 1} is derived from features refined by adapters, \textbf{pred 2} includes cache model knowledge, and \textbf{pred 3} is from linear probing.}
    \label{fig:feature}
    \vspace{-8pt}
\end{figure}

\section{Feature-Based Methods}
\label{sec:feature}
The prompt-based methods in \cref{sec:prompt} and parameter-based methods in \cref{sec:parameter} require forwards and backwards of the whole VLM to compute gradients, bringing high computation expenses and training time. Observing that features extracted by pretrained VLMs are discriminative and informative, feature-based methods aim to perform adaptation and refinements on the feature level for better efficiencies. In this paper, feature-based methods are categorized as \textit{cache-model-based} methods and \textit{refinement-based} methods. As illustrated in \cref{fig:feature}, feature refinement is achieved by adapters~\cite{houlsby2019parameter} or linear-probing~\cite{clip}. Cache models store representative features of input samples and refine the final outputs. Recent advances in both fields are summarized in \cref{tab:feature}.

\subsection{Feature Refinement}
\label{sec:feature_refine}
Rather than considering the whole VLM, feature refinement methods build upon the VLM-extracted vision and text features for further adaptation. These methods freeze the whole VLM and introduces external modules for various purposes, providing more dynamic adaptations.

\subsubsection{Refine with Adapters}
\label{sec:feature_refine_adapter}
A typical refinement framework is Adapters proposed in \cite{houlsby2019parameter}, where a lightweight linear adapter module $\Phi$ generates feature-dependent residuals for complement, as shown in \cref{eq:adapter}. Gao \textit{et al.}~\cite{gao2024clip} first adapts this scheme to VLMs by introducing dual adapters for each modality branch. Follow-up researches improve this paradigm by modifying the adapters' structures, positions, purposes and training objectives according to different adaptation settings. 

\begin{table*}[!t]
    \caption{Feature-based transfer methods for VLMs. Ca, A, TF represent cache-based, adapter-based and train-free methods.}
    \centering
    \label{tab:feature}
    \vspace{-8pt}
    \resizebox{\linewidth}{!}{
    \setlength{\tabcolsep}{4.7pt}
    \begin{tabular}{@{}p{2.06cm}cccl@{}}
    \toprule
    \textbf{Method} & \textbf{Venue} & \textbf{Setting} & \textbf{Type} & \textbf{Brief description} \\ \midrule
    CALIP~\cite{guo2023calip} & AAAI'23 & TTA & A & Promotes interactions between visual and textual features  using   parameter-free attention layers. \\
    MTA~\cite{zanella2024test} & CVPR'24 & TTA & A & Learns   quality assessment variables to assess the quality of each augmented view. \\
    VDPG~\cite{chi2024adapting} & ICLR'24 & TTA & A & Trains  a domain prompt generator to condense knowledge bank into domain-specific  prompts. \\
    L2C~\cite{chi2025learning} & ICLR'25 & TTA & A & Learns   directly on the input space with an independent side branch via invert   attention. \\
    PromptStyler~\cite{cho2023promptstyler} & ICCV'23 & DG & A & Learns   diverse prompt vectors of the same class for training linear classifier. \\
    GESTUR~\cite{lew2023gradient} & ICCVW'23 & DG & A & Estimates   unobservable gradients that reduces potential risks with parameterized   experts. \\
    CLIPCEIL~\cite{yu2024clipceil} & NeurIPS'24 & DG & A & Refines   the visual feature channels  for domain-invariance and   class-relevance with an adapter. \\
    Li \textit{et al.}~\cite{li2024prompt} & CVPR'24 & DG & A & Learns   a gating module to refine vision features with masks for both classification   and detection. \\
    PromptTA~\cite{zhang2025promptta} & ICASSP'25 & DG & A & Trains   a text adapter to store domain information and a classifier for prediction. \\
    TaskRes~\cite{yu2023task} & CVPR'23 & FSL & A & Tunes   a set of prior-independent parameters as a residual for encoding   task-specific knowledge. \\
    SHIP~\cite{wang2023improving} & ICCV'23 & FSL & A & Reconstructs   the visual features with synthesized prompts by variational autoencoders. \\ 
    CLAP~\cite{silva2024closer} & CVPR'24 & FSL & A & A   hyperparameter-tuning-free method with linear probing constrained by Lagrangian method. \\
    CLIPAdapter~\cite{gao2024clip} & IJCV'24 & FSL & A & Proposes   to apply adapters to vision or language branch for feature refinement. \\
    Candle~\cite{shi2024efficient} & KDD'24 & FSL & A & Proposes   compensating logit-adjusted loss for long-tail generalization with cross-modal attention. \\
    UniMoS~\cite{unimos} & CVPR'24 & UDA & A & Disentangles   VLMs' feature into vision and language-associated components for finer adaptation. \\
    Lai \textit{et al.}~\cite{lai2024empowering} & WACV'24 & UDA & A & Refines   vision and text features with independent parameters for   pseudo-labeling-based UDA \\
    PODA~\cite{fahes2023poda} & ICCV'23 & ZSL & A & Obtains   stylic features with different prompts then learns domain-wise segmenter   models. \\
    ULDA~\cite{yang2024unified} & CVPR'24 & ZSL & A & Aligns   pixel-level, regional-level, and scene-level features with text embeddings. \\
    LanDA~\cite{wang2024landa} & Arxiv & MSDA & A & Learns   domain-specific augmenters for image-free MSDA with only target textual   descriptions. \\
    ReCLIP~\cite{hu2024reclip} & WACV'24 & SFDA & A & Learns   a projection space to align visual and text features then update layer norms   of encoders. \\
    DAC~\cite{gondal2024domain} & WACV'24 & FSL & A,Ca & Introduces text, visual caches and a visual adapter for inter- and intra-modal alignment. \\
    CPL~\cite{zhang2024concept} & AAAI'24 & FSL & A,Ca & Creates a visual concept cache and a projector for vision-to-language prompting.\\
    DPE~\cite{zhang2024dual_proto} & NeurIPS'24 & TTA & Ca & Accumulates   task-specific knowledge with evolving textual and visual prototypes. \\
    TDA~\cite{tda} & CVPR'24 & TTA & Ca & Builds   key-value cache that stores few-shot pseudo labels and  corresponding features. \\
    DMN~\cite{zhang2024dual} & CVPR'24 & TTA & Ca & Builds dynamic and static memory for train-free zero- and few-shot  adaptation. \\
    BCA~\cite{zhou2025bayesian} & CVPR'25 & TTA & Ca & Updates   class embeddings to adapt likelihood while  updating the prior of class   embeddings. \\
    TipAdapter~\cite{zhang2022tip} & ECCV'22 & FSL & Ca & Builds   non-parametric key-value cache from few-shot training sets. \\
    SuS-X~\cite{udandarao2023sus} & ICCV'23 & FSL & Ca & Builds  cache models with support set generated by Stable Diffusion or retrieved   from datasets. \\
    APE~\cite{zhu2023not} & ICCV'23 & FSL & Ca & Refines   CLIP-extracted feature channels when building cache models. \\
    Wang \textit{et al.}~\cite{wang2024hard} & ICLR'24 & FSL & Ca & Builds   a Gaussian Discriminate Analysis classifier to complement  VLMs' zero-shot classifier. \\
    CuPL~\cite{pratt2023does} & ICCV'23 & FSL & TF & Utilizes   LLMs to construct more diverse category descriptions with more details. \\
    ZERO~\cite{farinafrustratingly} & NeurIPS'24 & TTA & TF & Augments   inputs, predicts then marginalizes over the most confident predictions. \\
    BendVLM~\cite{gerych2024bendvlm} & NeurIPS'24 & TTA & TF & A   nonlinear, train-free approach for TTA by VLM embedding debiasing. \\\bottomrule
    \end{tabular}}
\end{table*}

Observing that multi-head attention~\cite{vaswani2017attention} fosters cross-modal information exchanges, several methods~\cite{chi2025learning,guo2023calip,shi2024efficient,li2024prompt,chi2024adapting} introduce attention layers as adapters for the purposes of multimodal information fusion~\cite{li2024prompt,chi2025learning,chi2024adapting} and interactions~\cite{guo2023calip,shi2024efficient}.  In contrast to attention modules, adapters can be simplified into learnable parameters and variables. For example, TaskRes~\cite{yu2023task} learns task residuals that directly add to extracted text features. MTA~\cite{zanella2024test} learns inlierness variables to assess reliability of augmented views for TTA. In addition to feature refinement, adapters are also used for disentangling and encompassing the domain and modality information in the multimodal features. VDPG~\cite{chi2024adapting} encompasses domain information, PromptTA~\cite{zhang2025promptta} learns style features, and UniMoS~\cite{unimos} disentangles modality information by learning adapters. CLIPCEIL~\cite{yu2024clipceil} refines visual channels for domain-invariance and class-relevance. ReCLIP~\cite{hu2024reclip} and DAC~\cite{gondal2024domain} project vision and text features for distribution alignment. Adapters can also reconstruct or augment the features for various style information~\cite{wang2023improving,wang2024landa}.

\subsubsection{Linear Probing}
\label{sec:feature_refine_lp}
As discussed in \cite{clip}, linear probing (LP) is a simple yet effective approach that directly tunes a linear classifier on extracted features for downstream predictions. It is pointed out that LP even surpass full fine-tuning on OOD data~\cite{kumar2022fine}. To inherent such merits, several researches have integrated LP into their method designs.

One direction is to linear probe with augmented style features. PromptStyler~\cite{cho2023promptstyler} learns diverse style word vectors to generate style-content features for training a domain generalizable linear classifier. Similar approach is adopted to generate scene styles (e.g., at night, under rain, in snow) for zero-shot adaptation of semantic segmentation~\cite{fahes2023poda,yang2024unified}, where segmenter models are further trained. On the other hand, CLAP~\cite{silva2024closer} retains the strong zero-shot abilities in VLMs and avoids prototype degradation during linear probing with Augmented Lagrangian Multiplier methods. Based on modality-disentangled representations, UniMoS~\cite{unimos} directly adopts LP for generating vision-modality outputs, and PromptTA~\cite{zhang2025promptta} adopts a linear classifier and text adapter trained with style features for inference.

\subsection{Cache Model}
\label{sec:feature_cache}
Adapter-based methods still requires training the parameterized modules. Inspired by cache models~\cite{khandelwal2019generalization}, Tip-Adapter~\cite{zhang2022tip} proposes building cache models with VLM-extracted features. Denote the matrix of text features from all $C$ classes as $\mathbf{W}=(\mu_1, ..., \mu_C), \mathbf{W} \in \mathbb{R}^{d \times C}$, where $d$ is feature dimension. Given an input image with vision feature $v \in \mathbb{R}^d$, the zero-shot prediction logit before normalization can be computed with dot product: $\hat{y} = v\mathbf{W}$. The cache model is built by recording all vision features of few-shot training samples as keys $\mathbf{F} = (v_1, ..., v_N)$, and their corresponding one-hot labels as values $\mathbf{L} = (y_1, ..., y_N)$, where $\mathbf{F} \in \mathbb{R}^{d \times N}, \mathbf{L} \in \mathbb{R}^{C \times N}$. Given vision feature $v$ of a tested sample, the cache model complements the zero-shot prediction results in the residual manner: 
\begin{align}
    \hat{y} = \alpha \psi (v\mathbf{F})\mathbf{L}^{\top } + v\mathbf{W}, 
\label{eq:cache}
\end{align}
where $\psi(x) = \exp (-\beta(1-x))$, and $\alpha, \beta$ are hyperparameters. The first term in \cref{eq:cache} retrieves classification information of trained samples weighted by feature similarities, storing the target data information in a train-free manner. 

\begin{table*}[!t]
    \caption{Widely-used datasets for evaluating VLMs. }
    \centering
    \label{tab:dataset}
    \vspace{-8pt}
    \resizebox{\linewidth}{!}{
    \setlength{\tabcolsep}{5pt}
    \begin{tabular}{@{}clcccl@{}}
    \toprule
    \textbf{Task} & \textbf{Dataset} & \textbf{Setting} & \textbf{\# Class} & \textbf{\# Sample} & \textbf{Brief   description} \\ \midrule
    \multirow{21}{*}{\begin{tabular}[c]{@{}c@{}}Image\\ classification\end{tabular}} & ImageNet~\cite{deng2009imagenet} & FSL,DG,TTA & 1000 & 1,281,167 & A foundation image   classification dataset. \\
     & ImageNet-A~\cite{hendrycks2021natural} & FSL,DG,TTA & 200 & 7,500 & A variant of ImageNet   with natural adversarial examples. \\
     & ImageNet-R~\cite{hendrycks2021many} & FSL,DG,TTA & 200 & 30,000 & A variant of ImageNet   with renditions like art and cartoons. \\
     & ImageNet-S~\cite{wang2019learning} & FSL,DG,TTA & 1000 & 50,889 & A variant of ImageNet   with black-and-white sketch images. \\
     & ImageNet-V2~\cite{recht2019imagenet} & FSL,DG,TTA & 1000 & 10,000 & A non-overlapping   variant with similar distribution of ImageNet \\
     & Caltech101~\cite{fei2004learning} & FSL,TTA & 101 & 9,146 & Contains regular   objects for recognition. \\
     & OxfordPets~\cite{parkhi2012cats} & FSL,TTA & 37 & 7,349 & Contains fine-grained   breed labels of cats and dogs. \\
     & StanfordCars~\cite{krause20133d} & FSL,TTA & 196 & 16,185 & Contains fine-grained   vehicle models. \\
     & Flowers102~\cite{nilsback2008automated} & FSL,TTA & 102 & 8,189 & Contains fine-grained   flower species. \\
     & Food101~\cite{bossard2014food} & FSL,TTA & 101 & 101,000 & Contains images of   different foods with noises. \\
     & FGVC-Aircraft~\cite{maji2013fine} & FSL,TTA & 100 & 10,000 & Includes fine-grained   annotation of aircraft models. \\
     & SUN397~\cite{xiao2010sun} & FSL,TTA & 397 & 39,700 & Vision dataset for   scene recognition and understanding. \\
     & DTD~\cite{cimpoi2014describing} & FSL,TTA & 47 & 5,760 & Includes images of   regular texture for recognition. \\
     & EuroSAT~\cite{helber2019eurosat} & FSL,TTA & 10 & 2,700 & Includes satellite   sensing images of land cover situations. \\
     & UCF101~\cite{soomro2012ucf101} & FSL,TTA & 101 & 13,320 & Action recognition   dataset collected from video clips. \\
     & PACS~\cite{li2017deeper} & DG & 7 & 9,991 & Includes images from   domains \textit{photo}, \textit{art}, \textit{cartoon} and \textit{sketch}. \\
     & VLCS~\cite{torralba2011unbiased} & DG & 5 & 10,729 & Includes images from  Caltech101, LabelMe, SUN09 and VOC2007. \\
     & TerraIncognita~\cite{beery2018recognition} & DG & 10 & 24,330 & Includes animal   images taken at different locations. \\
     & OfficeHome~\cite{venkateswara2017deep} & DG,UDA & 65 & 15,500 & Includes office items   from domains \textit{Art}, \textit{Clipart}, \textit{Product}, \textit{RealWorld}. \\
     & DomainNet~\cite{peng2019moment} & DG,UDA & 345 & 586,575 & Includes pictures   from 6 domains for multi-source DA. \\
     & VisDA~\cite{peng2017visda} & UDA & 12 & 207,785 & Include synthetic and   real images for knowledge transfer. \\\midrule
    \multirow{3}{*}{\begin{tabular}[c]{@{}c@{}}Image-\\ text\end{tabular}} & VQAv2~\cite{goyal2017making} & QA & - & 265,016 & Includes images with   at least 3 open-ended questions to answer. \\
     & COCO caption~\cite{chen2015microsoft} & Retrieval & - & 82,783 & Includes images with   human-generated captions. \\
     & Flickr30k~\cite{young2014image} & Retrieval & - & 31,783 & Includes images with   sentence-based image description. \\\midrule
    \multirow{3}{*}{\begin{tabular}[c]{@{}c@{}}Video-\\ text\end{tabular}} & TVQA~\cite{lei2018tvqa} & QA & - & 21,793 & Includes QA pairs   from 6 TV shows and 21793 video clips. \\
     & How2QA~\cite{li2020hero} & QA & - & 9,035 & Includes QA pairs   with one correct answer and three distractors. \\
     & TVC~\cite{lei2020tvr} & Caption & - & 108,965 & Includes short video   moments and corresponding captions. \\\midrule
    \multirow{2}{*}{Navigation} & R2R~\cite{anderson2018vision} & VLN & - & 21,567 & Includes building   scenes with language instructions for navigation. \\
     & REVERIE~\cite{qi2020reverie} & VLN & - & 21,702 & An indoor grounding   dataset with target objects and instructions. \\\midrule
    \multirow{2}{*}{\begin{tabular}[c]{@{}c@{}}Object \\ detection\end{tabular}} & COCO 2014~\cite{lin2014microsoft} & OD & 80 & 83,000 & \multirow{2}{*}{\begin{tabular}[c]{@{}l@{}}Large-scale   datasets for object detection, segmentation, keypoint \\ detection, and   captioning.\end{tabular}} \\
     & COCO 2017~\cite{lin2014microsoft} & OD & 80 & 118,000 &  \\\midrule
    \multirow{2}{*}{\begin{tabular}[c]{@{}c@{}}Semantic \\ segmentation\end{tabular}} & Cityscapes~\cite{cordts2016cityscapes} & SS & 19 & 2,975 & Focuses on semantic   understanding of urban street scenes. \\
     & GTA5~\cite{richter2016playing} & SS & 19 & 24,966 & Pixel-level semantic   annotations from the game Grand Theft Auto 5. \\ \bottomrule
    \end{tabular}}
\end{table*}

Relevant works aim to improve this framework. SuS-X~\cite{udandarao2023sus} enhances the vision-modality cache model in \cref{eq:cache} by incorporating both vision and text features. SuS-X first computes cross-modal similarities between cache values and text features: $\mathbf{S} = \text{Softmax} (\mathbf{F}\mathbf{W}^{\top})$, as well as normalized zero-shot probabilities of test samples: $s = \text{Softmax}(v\mathbf{W})$. An affinity matrix $\mathbf{M}$ can be computed by computing KL divergence: $M_{i,j} = \text{KL}(s_i, \mathbf{S}_j)$. Finally, prediction result in \cref{eq:cache} is modified as:
\begin{align}
    \hat{y} = \alpha \psi (v\mathbf{F})\mathbf{L}^{\top } + v\mathbf{W} + \gamma \omega (-\mathbf{M})\mathbf{L},
\label{eq:sus-x}
\end{align}
where $\gamma$ is hyperparameter and $\omega(\cdot)$ rescales the values in $\mathbf{M}$. The third term in \cref{eq:sus-x} introduces vision-language similarities when retrieving from the cache. APE~\cite{zhu2023not} first refines the features in $\mathbf{F},v,\mathbf{W}$ by selecting $d'$ most informative channels from all $d$ feature dimensions, obtaining $\mathbf{F}',v',\mathbf{W}'$. Then, the prediction in \cref{eq:cache} is modified as:
\begin{align}
    \hat{y} = \alpha \psi (v'\mathbf{F}')(\text{diag}(\mathbf{R'}) \mathbf{L}) + v\mathbf{W},
\label{eq:ape}
\end{align}
where $\mathbf{R'} = \exp(\gamma \text{KL}(\mathbf{L}, \mathbf{F}'\mathbf{W}'^\top))$. Compared with \cref{eq:cache}, the first term in \cref{eq:ape} introduces feature informativeness measured by the refined feature channels. Wang \textit{et al.}~\cite{wang2024hard} further improves the cache model by estimating weight and bias of the classifier using Gaussian Discriminate Analysis.

While the cache-model-based methods introduced above are designed for few-shot learning (FSL) with labeled samples, their high efficiency brought by the train-free design makes them extremely suitable for TTA. TDA~\cite{tda} modifies \cref{eq:cache} by constructing the key matrix $\mathbf{F}$ with high quality test samples, whose highly-confident pseudo labels are used for the value matrix $\mathbf{L}$. A negative cache is also built with low-quality samples to further enhance TTA. To simultaneously handle test-time and few-shot adaptation settings, DMN~\cite{zhang2024dual} introduces a dynamic cache to store test-time samples and a static cache for optional few-shot training samples. Rather than building cache with extracted features, these representative prototypes can also be learned as the test-time adaptation proceeds~\cite{zhang2024dual_proto,zhou2025bayesian}. Cache models can be integrated with adapters that further refine the cache contents~\cite{gondal2024domain,zhang2024concept}. In addition to cache models, there are other train-free (TF) methods aided by external knowledge~\cite{pratt2023does}, utilizing various augmentations~\cite{farinafrustratingly} or embedding debiasing~\cite{gerych2024bendvlm}, as summarized in \cref{tab:feature}.

\subsection{Discussion}
Feature-based methods exploit the strong representation ability of pretrained VLMs by performing adaptation mainly on extracted features. Refinement-based methods in \cref{sec:feature_refine} introduces external adaptation modules like Adapters~\cite{houlsby2019parameter} to modify the features, or replace the cross-modal prediction head (\cref{eq:inference}) by directly training a linear classifier on vision features. Some researches have pointed out that  VLMs' pretrained features sometimes surpasses the fine-tuned features~\cite{kumar2022fine,silva2024closer}, which motivates train-free cache-model-based methods. These methods record the extracted features of train samples as class prototypes to benefit the zero-shot prediction. To conclude, feature-based methods provide more computation efficiency, can are suitable for agile adaptation and deployment of VLMs in resource-constrained scenarios like edge devices.

\section{Datasets and Experimental Results}
\label{sec:experiment}
This section reviews popular datasets used for evaluating the generalization of VLMs and comprehensively compares the performances of existing knowledge transfer methods.

\begin{table}[!t]
    \caption{Performance comparison on domain-level DG tasks.}
    \centering
    \label{tab:result_2}
    \vspace{-8pt}
    \resizebox{\linewidth}{!}{
    \setlength{\tabcolsep}{4pt}
    \begin{tabular}{@{}lcc|cccccc@{}}
    \toprule
    \textbf{Method} & \textbf{Type} & \textbf{Backbone} & PACS & VLCS & OH & DN & TI & \textbf{Avg.} \\ \midrule
    DFKD-VLFM~\cite{xuan2023distilling}  & KD & ResNet50 & 78.3 & 76.0 & - & - & - & - \\ 
    RISE~\cite{huang2023sentence}  & KD & ResNet50 & 90.2 & 82.4 & 72.6 & - & 54.0 & - \\
    DART~\cite{jain2023dart}  & FT-R & ResNet50 & 88.9 & 80.3 & 71.9 & 47.1 & 51.3 & 67.9 \\\midrule
    CLIP-zeroshot~\cite{clip} & - & ViT-B/16 & 96.2 & 81.7 & 82.0 & 57.5 & 33.4 & 70.2  \\
    VL2V-ADiP~\cite{addepalli2024leveraging}  & KD & ViT-B/16 & 96.7 & 83.3 & 87.4 & 62.8 & 58.5 & 77.7 \\
    STYLIP~\cite{bose2024stylip}  & T & ViT-B/16 & 97.0 & 82.9 & 83.9 & 68.6 & - & - \\
    SPG~\cite{bai2024soft}  & T & ViT-B/16 & 97.0 & 82.4 & 83.6 & 60.1 & 50.2 & 74.7 \\
    DSPL~\cite{cheng2024disentangled}  & V,T & ViT-B/16 & 97.5 & 86.4 & 86.1 & 62.1 & 57.1 & 77.8 \\
    Xiao \textit{et al.}~\cite{xiao2024any}  & V,T & ViT-B/16 & 98.5 & 87.0 & 86.0 & 61.8 & - & - \\
    SWAD~\cite{swad}  & FT-R & ViT-B/16 & 91.3 & 79.4 & 76.9 & 51.7 & 45.4 & 68.9 \\
    MIRO~\cite{miro}  & FT-R & ViT-B/16 & 95.6 & 82.2 & 82.5 & 54.0 & 54.3 & 73.7 \\
    CLIPood~\cite{clipood}  & FT-R & ViT-B/16 & 97.3 & 85.0 & 87.0 & 63.5 & 60.4 & 78.6 \\
    UEO~\cite{ueo}  & FT-R & ViT-B/16 & 96.9 & 81.3 & 86.0 & 60.8 & 51.5 & 75.3 \\
    CAR-FT~\cite{mao2024context}  & FT-R & ViT-B/16 & 96.8 & 85.5 & 85.7 & 62.5 & 61.9 & 78.5 \\
    DPL~\cite{zhang2023domain}  & A & ViT-B/16 & 96.4 & 80.9 & 83.0 & 59.5 & 46.6 & 73.3 \\
    GESTUR~\cite{lew2023gradient}  & A & ViT-B/16 & 96.0 & 82.8 & 84.2 & 58.9 & 55.7 & 75.5 \\
    CLIPCEIL~\cite{yu2024clipceil}  & A & ViT-B/16 & 97.6 & 88.4 & 85.4 & 62.0 & 53.0 & 77.3 \\
    PromptStyler~\cite{cho2023promptstyler}  & A & ViT-B/16 & 97.2 & 82.9 & 83.6 & 59.4 & - & - \\
    PromptTA~\cite{zhang2025promptta}  & A & ViT-B/16 & 97.3 & 83.6 & 82.9 & 59.4 & - & - \\\bottomrule
    \end{tabular}}
    \vspace{-8pt}
\end{table}

\subsection{Datasets}
\label{sec:experiment_dataset}
\cref{tab:dataset} summarizes representative datasets on image, text, video modalities for tasks including classification, retrieval, captioning, question answering (QA), vision-language navigation (VLN), object detection (OD) and semantic segmentation (SS). The \textbf{Setting} column indicates the problems that the dataset is designed to solve.

\begin{table*}[!t]
    \caption{Performance comparison over various methods, settings and backbones on dataset-level generalization. Method types correspond with the types introduced in \cref{tab:prompt}, \cref{tab:parameter} and \cref{tab:feature}. All methods are based on CLIP~\cite{clip} with different vision backbones.}
    \centering
    \label{tab:result_1}
    \vspace{-5pt}
    \resizebox{\linewidth}{!}{
    \setlength{\tabcolsep}{4.7pt}
    \begin{tabular}{@{}lcc|cccccc|ccccccccccc@{}}
    \toprule
    \multirow{2}{*}{\textbf{Method}} & \multirow{2}{*}{\textbf{Type}} & \multirow{2}{*}{\textbf{Backbone}} & \multicolumn{6}{c|}{\textbf{ImageNet family}} & \multicolumn{11}{c}{\textbf{Few-shot   datasets}} \\ \cmidrule(l){4-20} 
     &  &  & -I & -V2 & -S & -A & -R & \textbf{Avg.} & Calt & Pets & Cars & Flower & Food & FGVC & SUN & DTD & Euro & UCF & \textbf{Avg.} \\\midrule
    \multicolumn{2}{@{}l}{CLIP-zeroshot~\cite{clip}} & ResNet50 & 58.2 & 51.4 & 33.3 & 21.7 & 56.0 & 44.1 & 85.1 & 83.1 & 55.7 & 65.4 & 74.2 & 17.1 & 58.6 & 42.2 & 37.6 & 61.2 & 58.0 \\
    \multicolumn{2}{@{}l}{CLIP-zeroshot~\cite{clip}} & ViT-B/16 & 66.7 & 60.8 & 46.2 & 47.8 & 74.0 & 59.1 & 93.6 & 86.9 & 66.1 & 67.0 & 82.9 & 23.2 & 65.6 & 45.0 & 50.4 & 65.2 & 64.6 \\
    \multicolumn{2}{@{}l}{CLIP-zeroshot~\cite{clip}} & ViT-B/16$^\dagger$ & - & - & - & - & - & - & 95.4 & 94.1 & 68.7 & 74.8 & 90.7 & 31.1 & 72.2 & 56.4 & 60.0 & 73.9 & 71.7 \\\midrule
    \multicolumn{20}{@{}l}{\cellcolor[HTML]{EFEFEF}\textit{Few-shot learning (FSL) methods (16-shot results are provided)}} \\
    VPT~\cite{vpt}  & V & ViT-B/16 & - & - & - & - & - & - & 96.4 & 96.8 & 73.1 & 81.1 & 91.6 & 34.7 & 78.5 & 67.3 & 77.7 & 79.0 & 77.6 \\
    DPT~\cite{xing2023dual}  & V,T & ViT-B/16 & - & - & - & - & - & - & 95.6 & 91.2 & 82.6 & 96.6 & 79.3 & 48.4 & 71.0 & 70.2 & 91.2 & 81.4 & 80.7 \\
    UPT~\cite{zang2022unified}  & V,T & ViT-B/16 & 72.6 & 64.4 & 48.7 & 50.7 & 76.2 & 62.5 & 95.9 & 93.0 & 84.3 & 97.1 & 85.0 & 46.8 & 75.9 & 70.7 & 90.5 & 84.0 & 82.3 \\
    Wang \textit{et al.}~\cite{wang2024hard}  & Ca & ViT-B/16 & - & - & - & - & - & - & 92.6 & 88.8 & 75.1 & 95.7 & 79.1 & 40.6 & 70.7 & 66.5 & 86.1 & 77.5 & 77.3 \\
    APE~\cite{zhu2023not}  & Ca & ViT-B/16 & - & - & - & - & - & - & 92.3 & 88.0 & 70.5 & 92.0 & 78.4 & 31.2 & 69.6 & 67.4 & 78.4 & 74.5 & 74.2 \\
    CLAP~\cite{silva2024closer}  & A & ViT-B/16 & 71.8 & 64.1 & 47.7 & 48.4 & 76.7 & 61.7 & - & - & - & - & - & - & - & - & - & - & - \\
    Candle~\cite{shi2024efficient}  & A & ViT-B/16 & 71.6 & 62.8 & 48.3 & 49.1 & 75.0 & 61.4 & 91.3 & 88.9 & 64.6 & 68.3 & 85.5 & 24.2 & 66.1 & 44.6 & 48.4 & 67.2 & 64.9 \\
    TaskRes~\cite{yu2023task}  & A & ViT-B/16 & 73.1 & 65.3 & 49.1 & 50.4 & 77.7 & 63.1 & 93.4 & 87.8 & 76.8 & 96.0 & 77.6 & 36.3 & 70.7 & 67.1 & 84.0 & 78.0 & 76.8 \\
    CoCoOp~\cite{cocoop}  & T & ViT-B/16$^\dagger$ & 71.0 & 64.1 & 48.8 & 50.6 & 76.2 & 62.1 & 95.8 & 96.4 & 72.0 & 81.7 & 91.0 & 27.7 & 78.3 & 64.9 & 71.2 & 77.6 & 75.7 \\
    CoOp~\cite{coop}  & T & ViT-B/16$^\dagger$ & 71.5 & 64.2 & 48.0 & 49.7 & 75.2 & 61.7 & 93.7 & 94.5 & 68.1 & 74.1 & 85.2 & 28.8 & 72.5 & 54.2 & 68.7 & 67.5 & 70.7 \\
    KgCoOp~\cite{yao2023visual}  & T & ViT-B/16$^\dagger$ & 71.2 & 64.1 & 49.0 & 50.7 & 76.7 & 62.3 & 96.0 & 96.2 & 73.4 & 83.7 & 91.1 & 34.8 & 78.4 & 64.4 & 73.5 & 79.7 & 77.1 \\
    ProGrad~\cite{zhu2023prompt}  & T & ViT-B/16$^\dagger$ & 70.5 & 63.4 & 48.2 & 49.5 & 75.2 & 61.3 & 95.9 & 96.3 & 72.9 & 82.0 & 90.0 & 32.8 & 77.6 & 62.5 & 72.7 & 79.4 & 76.2 \\
    DePT~\cite{zhang2024dept}  & T & ViT-B/16$^\dagger$ & - & - & - & - & - & - & 96.3 & 96.4 & 77.8 & 86.5 & 91.2 & 40.7 & 81.1 & 71.1 & 84.9 & 82.3 & 80.8 \\
    DeCoOp~\cite{zhou2024decoop}  & T & ViT-B/16$^\dagger$ & - & - & - & - & - & - & 96.5 & 95.3 & 73.2 & 84.2 & 90.7 & 31.4 & 78.1 & 62.7 & 74.6 & 77.7 & 76.4 \\
    DPC~\cite{li2025dpc}  & T & ViT-B/16$^\dagger$ & - & - & - & - & - & - & 96.5 & 96.7 & 78.5 & 83.6 & 91.5 & 40.2 & 80.8 & 66.8 & 84.1 & 83.2 & 80.2 \\
    MaPLe~\cite{maple}  & C & ViT-B/16$^\dagger$ & 70.7 & 64.1 & 49.2 & 50.9 & 77.0 & 62.4 & 96.0 & 96.6 & 73.5 & 82.6 & 91.4 & 36.5 & 79.8 & 68.2 & 82.4 & 80.8 & 78.8 \\
    ZIP~\cite{park2025zip}  & C & ViT-B/16$^\dagger$ & 66.2 & 59.7 & 45.5 & 47.1 & 75.2 & 58.7 & 95.6 & 95.9 & - & 72.7 & 89.9 & 31.2 & 70.9 & 55.8 & 72.9 & 72.2 & - \\
    PromptSRC~\cite{khattak2023self}  & V,T & ViT-B/16$^\dagger$ & 71.3 & 64.4 & 49.6 & 50.9 & 77.8 & 62.8 & 96.0 & 96.3 & 76.6 & 86.0 & 91.1 & 40.2 & 80.5 & 71.8 & 82.3 & 82.7 & 80.3 \\
    DAPT~\cite{cho2023distribution}  & V,T & ViT-B/16$^\dagger$ & 72.2 & 64.9 & 48.3 & 48.7 & 75.8 & 62.0 & 92.7 & 81.8 & 66.8 & 75.5 & 89.7 & 27.5 & 78.2 & 64.7 & 59.2 & 77.7 & 71.4 \\
    ProText~\cite{khattak2025learning}  & C,T & ViT-B/16$^\dagger$ & 70.2 & 63.5 & 49.5 & 51.5 & 77.4 & 62.4 & 96.8 & 96.5 & 69.8 & 76.4 & 91.1 & 32.4 & 77.6 & 62.3 & 68.7 & 77.5 & 74.9 \\
    CuPL~\cite{pratt2023does}  & TF & ViT-B/16$^\dagger$ & 69.6 & 63.3 & 49.0 & 50.7 & 77.1 & 61.9 & 95.6 & 95.8 & 68.8 & 76.1 & 90.6 & 33.2 & 75.7 & 61.3 & 61.8 & 76.4 & 73.5 \\
    CPL~\cite{zhang2024concept}  & A,Ca & ViT-B/16$^\dagger$ & - & - & - & - & - & - & 96.7 & 97.0 & 78.0 & 88.4 & 92.9 & 40.5 & 80.8 & 70.4 & 87.1 & 83.3 & 81.5 \\
    SHIP~\cite{wang2023improving}  & A & ViT-B/16$^\dagger$ & - & - & - & - & - & - & 96.3 & 96.0 & 77.2 & 85.7 & 91.1 & 39.0 & 79.4 & 70.2 & 87.1 & 81.9 & 80.4 \\\midrule
    \multicolumn{20}{@{}l}{\cellcolor[HTML]{EFEFEF}\textit{Domain generalization (DG) methods}} \\
    KDPL~\cite{mistretta2024improving}  & KD &  ViT-B/32 & 66.3 & 58.6 & 43.1 & 31.7 & 68.8 & 53.7 & 93.6 & 88.2 & 60.4 & 65.1 & 80.9 & 18.2 & 65.7 & 43.3 & 44.3 & 63.4 & 62.3 \\
    OGEN~\cite{zang2024overcoming}  & T & ViT-B/16 & - & - & - & - & - & - & 95.1 & 91.4 & 66.0 & 72.9 & 86.5 & 23.0 & 68.4 & 46.4 & 45.8 & 69.7 & 66.5 \\
    STYLIP~\cite{bose2024stylip}  & T & ViT-B/16 & - & - & - & - & - & - & 95.5 & 91.6 & 67.1 & 72.4 & 88.6 & 25.2 & 68.1 & 47.9 & 48.2 & 69.3 & 67.4 \\
    Hao  \textit{et al.}~\cite{hao2024quantized}  & C & ViT-B/16 & 70.7 & 63.9 & 48.9 & 51.1 & 76.9 & 62.3 & 94.1 & 90.5 & 66.0 & 71.3 & 86.2 & 22.7 & 66.8 & 44.2 & 48.2 & 69.2 & 65.9 \\
    CLIPood~\cite{zhang2024amend}  & FT-R & ViT-B/16 & 71.6 & 64.9 & 49.3 & 50.4 & 77.2 & 62.7 & - & - & - & - & - & - & - & - & - & - & - \\
    UEO~\cite{ueo}  & FT-R & ViT-B/16 & 71.9 & 65.1 & 48.6 & 48.6 & 75.9 & 62.0 & - & - & - & - & - & - & - & - & - & - & - \\
    CaRot~\cite{oh2024towards}  & FT-R & ViT-B/16 & 83.1 & 74.1 & 52.7 & 51.6 & 77.7 & 67.9 & - & - & - & - & - & - & - & - & - & - & - \\
    LipsumFT~\cite{nam2024lipsum}  & FT-R & ViT-B/16 & 83.3 & 73.6 & 51.4 & 49.9 & 75.9 & 66.8 & - & - & - & - & - & - & - & - & - & - & - \\
    CAR-FT~\cite{mao2024context}  & FT-R & ViT-B/16 & 83.8 & 74.0 & 53.0 & 49.5 & 75.4 & 67.1 & - & - & - & - & - & - & - & - & - & - & - \\
    SAFT~\cite{nguyen2024saft}  & FT-S & ViT-B/16 & 72.7 & 65.8 & 49.7 & 51.6 & 78.1 & 63.6 & 94.2 & 90.9 & 65.3 & 70.8 & 86.3 & 25.2 & 68.2 & 46.2 & 49.6 & 70.2 & 66.7 \\
    PromptKD~\cite{li2024promptkd}  & KD &  ViT-B/16 & - & - & - & - & - & - & 93.6 & 91.6 & 73.9 & 75.3 & 88.8 & 26.2 & 68.6 & 55.1 & 63.7 & 76.4 & 71.3 \\
    APD~\cite{luo2024adversarial}  & KD &  ViT-B/16 & - & - & - & - & - & - & 93.9 & 88.5 & 81.1 & 94.4 & 71.5 & 47.8 & 71.4 & 65.5 & 76.9 & 77.1 & 76.8 \\\midrule
    \multicolumn{20}{@{}l}{\cellcolor[HTML]{EFEFEF}\textit{Test-time adaptation (TTA) methods}} \\
    Ma \textit{et al.}~\cite{ma2023swapprompt}  & T & ResNet50 & 61.8 & 53.9 & 38.2 & 24.5 & 60.9 & 47.9 & 89.9 & 89.1 & 59.6 & 70.2 & 75.1 & - & 63.9 & 47.3 & 46.6 & 65.7 & - \\
    CALIP~\cite{guo2023calip}  & A & ResNet50 & - & - & - & - & - & - & 87.7 & 86.2 & 56.3 & 66.4 & 77.4 & 17.8 & 58.6 & 42.4 & 38.9 & 61.7 & 59.3 \\
    R-TPT~\cite{sheng2025r}  & T & ResNet50 & - & - & - & - & - & - & 86.7 & 84.6 & 58.1 & 60.6 & - & 17.5 & - & 41.3 & 21.2 & 59.7 & - \\
    DiffTPT~\cite{difftpt}  & T & ViT-B/16 & 70.3 & 65.1 & 46.8 & 55.7 & 75.0 & 62.6 & 92.5 & 88.2 & 67.0 & 70.1 & 87.2 & 25.6 & 65.7 & 47.0 & 43.1 & 68.2 & 65.5 \\
    C-TPT~\cite{yoon2024c}  & T & ViT-B/16 & 69.3 & 63.4 & 48.5 & 52.9 & 78.0 & 62.4 & 94.1 & 87.4 & 66.7 & 69.9 & 84.5 & 23.9 & 66.0 & 46.8 & 48.7 & 66.7 & 65.5 \\
    TPT~\cite{tpt}  & T & ViT-B/16 & 73.6 & 66.8 & 49.3 & 58.0 & 77.3 & 65.0 & 94.2 & 87.8 & 66.9 & 69.0 & 84.7 & 24.8 & 65.5 & 47.8 & 42.4 & 68.0 & 65.1 \\
    O-TPT~\cite{sharifdeen2025tpt}  & T & ViT-B/16 & 67.3 & 61.7 & 47.1 & 49.9 & 72.6 & 59.7 & 94.0 & 88.0 & 64.5 & 70.1 & 84.1 & 23.6 & 64.2 & 45.7 & 42.8 & 64.2 & 64.1 \\ 
    Abdul \textit{et al.}~\cite{abdul2023align}  & T & ViT-B/16 & - & 65.3 & 50.2 & 59.4 & 79.3 & 63.6 & 94.0 & 90.8 & 68.5 & 72.4 & 86.7 & 24.8 & 67.5 & 47.2 & 47.9 & 69.5 & 66.9 \\
    Xiao \textit{et al.}~\cite{xiao2025dynaprompt}  & T & ViT-B/16 & 69.6 & 64.7 & 48.2 & 56.2 & 78.2 & 63.4 & 94.3 & 88.3 & 67.7 & 70.0 & 85.4 & 24.3 & 66.3 & 48.0 & 42.3 & 68.7 & 65.5 \\
    WATT~\cite{watt}  & FT-S & ViT-B/16 & 64.1 & 62.4 & 49.0 & 51.0 & 75.4 & 60.4 & 93.6 & 88.1 & 66.8 & 68.5 & 84.8 & 24.2 & 65.8 & 46.9 & 52.0 & 65.7 & 65.6 \\
    TTL~\cite{imam2025test}  & FT-S & ViT-B/16 & 70.2 & 64.6 & 48.6 & 60.5 & 77.5 & 64.3 & 93.6 & 88.7 & 68.0 & 70.5 & 85.1 & 23.8 & 66.3 & 46.7 & 42.0 & 69.2 & 65.4 \\
    DPE~\cite{zhang2024dual_proto}  & Ca & ViT-B/16 & 71.9 & 65.4 & 52.3 & 59.6 & 80.4 & 65.9 & 94.8 & 91.1 & 67.3 & 75.1 & 86.2 & 29.0 & 70.1 & 54.2 & 55.8 & 70.4 & 69.4 \\
    TDA~\cite{tda}  & Ca & ViT-B/16 & 69.5 & 64.7 & 50.5 & 60.1 & 80.2 & 65.0 & 94.2 & 88.6 & 67.3 & 71.4 & 86.1 & 23.9 & 67.6 & 47.4 & 58.0 & 70.7 & 67.5 \\
    DMN~\cite{zhang2024dual}  & Ca & ViT-B/16 & 72.3 & 65.2 & 53.2 & 58.3 & 78.6 & 65.5 & 95.4 & 92.0 & 68.0 & 74.5 & 85.1 & 30.0 & 70.2 & 55.9 & 59.4 & 72.5 & 70.3 \\
    BCA~\cite{zhou2025bayesian}  & Ca & ViT-B/16 & 70.2 & 64.9 & 50.9 & 61.1 & 80.7 & 65.6 & 94.7 & 90.4 & 66.9 & 73.1 & 86.0 & 28.6 & 68.4 & 53.5 & 56.6 & 67.6 & 68.6 \\
    MTA~\cite{zanella2024test}  & A & ViT-B/16 & 70.1 & 64.2 & 49.6 & 58.1 & 78.3 & 64.1 & 94.2 & 88.2 & 68.5 & 68.1 & 85.0 & 25.2 & 66.7 & 45.9 & 45.4 & 68.7 & 65.6 \\
    HisTPT~\cite{zhang2024historical}  & T,Ca & ViT-B/16 & - & - & - & - & - & - & 94.5 & 89.1 & 69.2 & 71.2 & 89.3 & 26.9 & 67.2 & 48.9 & 49.7 & 70.1 & 67.6 \\
    RLCF~\cite{zhao2024test}  & KD & ViT-B/16 & 75.5 & 70.4 & 57.7 & 75.2 & 87.2 & 73.2 & - & - & - & - & - & - & - & - & - & - & - \\
    ZERO~\cite{farinafrustratingly}  & TF & ViT-B/16 & 70.9 & 65.2 & 50.3 & 64.1 & 80.8 & 66.2 & 94.1 & 87.2 & 68.5 & 66.8 & 84.6 & 24.4 & 66.9 & 45.9 & - & 68.6 & - \\\bottomrule
    \multicolumn{20}{@{}l}{$^\dagger$ applies harmonic mean (HM) on few-shot datasets, as defined in~\cite{xian2017zero}.}
    \end{tabular}}
    \vspace{-5pt}
\end{table*}

We can observe that the majority of works focus on image classification, where the VLM needs to generalize to shifted data distributions~\cite{venkateswara2017deep,peng2019moment,hendrycks2021natural,hendrycks2021many,wang2019learning}, fine-grained categories~\cite{parkhi2012cats,krause20133d,nilsback2008automated,maji2013fine} or specialized data not included in VLMs' pretraining datasets~\cite{cimpoi2014describing,helber2019eurosat,maji2013fine}. With slight modifications, their vision perception abilities can be applied to more visual tasks like object detection~\cite{lin2014microsoft} and semantic segmentation~\cite{cordts2016cityscapes,richter2016playing}. In addition, the vision-language processing ability of VLMs enables them to solve diverse multimodal tasks including visual question answering~\cite{goyal2017making,lei2018tvqa,li2020hero}, image-text retrieval~\cite{chen2015microsoft,young2014image}, and vision-language navigation~\cite{anderson2018vision} where an agent is expected to complete navigation tasks as instructed by natural languages~\cite{anderson2018vision,anderson2018vision}.

\subsection{Experimental Results}
\label{sec:experiment_results}
Aligning with recent research focus, this survey mainly provides performance comparisons for \textit{image classification} tasks evaluated by classification accuracies. The comparison covers various generalization settings introduced in \cref{sec:background_setting}. The compared methods adopt CLIP~\cite{clip} with different backbones in the vision encoder, which are indicated in the \textbf{Backbone} column of each result table. The method types summarized in \cref{tab:prompt}, \cref{tab:feature}, \cref{tab:parameter} are also included.

\begin{table*}[!t]
    \caption{Performance comparison of domain-wise transfer tasks. On OfficeHome, all 12 sets of cross-domain results are provided, e.g., task \textit{AC} means adaptation from source domain \textit{A} to target domain \textit{C}. On other datasets the averaged results are provided.}
    \centering
    \label{tab:result_3}
    \vspace{-8pt}
    \resizebox{\linewidth}{!}{
    \setlength{\tabcolsep}{4.7pt}
    \begin{tabular}{@{}lc|cccccccccccccccccc@{}}
    \toprule
    \multirow{2}{*}{\textbf{Method}}  & \multirow{2}{*}{\textbf{Type}} & \multicolumn{14}{c}{\textbf{OfficeHome}} & \multicolumn{2}{c}{\textbf{VisDA}} & \multicolumn{2}{c}{\textbf{DomainNet}} \\ \cmidrule(l){3-20} 
     & & \textbf{Backbone} & AC & AP & AR & CA & CP & CR & PA & PC & PR & RA & RC & RP & \textbf{Avg.} & \textbf{Backbone} & \textbf{Avg.} & \textbf{Backbone} & \textbf{Avg.} \\ \cmidrule(r){1-2} \cmidrule(l){3-16} \cmidrule(l){17-18} \cmidrule(l){19-20}
    \multicolumn{2}{@{}l|}{CLIP-zeroshot~\cite{clip}} & ResNet50 & 51.7 & 81.5 & 82.3 & 71.7 & 81.5 & 82.3 & 71.7 & 51.7 & 82.3 & 71.7 & 51.7 & 81.5 & 71.8 & ResNet101 & 84.4 & ViT-B/16 & 56.2 \\
    \multicolumn{2}{@{}l|}{CLIP-zeroshot~\cite{clip}} & ViT-B/16 & 67.8 & 89.0 & 89.8 & 82.9 & 89.0 & 89.8 & 82.9 & 67.8 & 89.8 & 82.9 & 67.8 & 89.0 & 82.4 & ViT-B/16 & 88.9 & - & - \\\midrule
    \multicolumn{20}{@{}l}{\cellcolor[HTML]{EFEFEF}\textit{Unsupervised domain adaptation (UDA) methods (sorted by mean accuracy on OfficeHome)}} \\
    Xiao \textit{et al.}~\cite{xiao2024any} & V,T & ResNet50 & 51.6 & 81.9 & 82.6 & 71.9 & 81.9 & 82.6 & 71.9 & 51.6 & 82.6 & 71.9 & 51.6 & 81.9 & 72.0 & ResNet101 & 84.4 & ViT-B/16 & 56.2 \\
    CLIP-Div~\cite{zhu2024clip}  & KD & ResNet50 & 57.2 & 80.4 & 82.9 & 73.9 & 80.7 & 81.1 & 72.8 & 58.6 & 83.5 & 73.3 & 59.9 & 81.9 & 73.9 & ResNet101 & 80.8 & ViT-B/16 & 54.6 \\
    DAPrompt~\cite{daprompt}  & T & ResNet50 & 54.1 & 84.3 & 84.8 & 74.4 & 83.7 & 85.0 & 74.5 & 54.6 & 84.8 & 75.2 & 54.7 & 83.8 & 74.5 & ResNet101 & 86.9 & ViT-B/16 & 59.8 \\
    Shi \textit{et al.}~\cite{shi2024clip}  & T & ResNet50 & 54.6 & 84.6 & 85.1 & 75.8 & 84.2 & 85.1 & 74.5 & 54.1 & 85.2 & 75.2 & 54.9 & 84.0 & 74.8 & ResNet101 & 86.8 & ViT-B/16 & 59.5 \\
    FUZZLE~\cite{shi2024unsupervised}  & T & ResNet50 & 55.9 & 84.6 & 85.5 & 75.0 & 84.8 & 84.3 & 74.0 & 56.2 & 85.1 & 75.1 & 56.2 & 86.4 & 75.3 & ResNet101 & 86.5 & ViT-B/16 & 59.5 \\
    PDA~\cite{bai2024prompt}  & V,T & ResNet50 & 55.4 & 85.1 & 85.8 & 75.2 & 85.2 & 85.2 & 74.2 & 55.2 & 85.8 & 74.7 & 55.8 & 86.3 & 75.3 & ResNet101 & 86.4 & - & - \\
    AD-CLIP~\cite{singha2023ad}  & T & ResNet50 & 55.4 & 85.2 & 85.6 & 76.1 & 85.8 & 86.2 & 76.7 & 56.1 & 85.4 & 76.8 & 56.1 & 85.5 & 75.9 & ResNet101 & 87.7 & - & - \\
    Lai \textit{et al.}~\cite{lai2024empowering}  & A & ResNet50 & 58.1 & 85.0 & 84.5 & 77.4 & 85.0 & 84.7 & 76.5 & 58.8 & 85.7 & 75.9 & 60.4 & 86.4 & 76.5 & ResNet101 & 89.2 & ViT-B/16 & 62.3 \\
    PADCLIP~\cite{lai2023padclip}  & FT-R & ResNet50 & 57.5 & 84.0 & 83.8 & 77.8 & 85.5 & 84.7 & 76.3 & 59.2 & 85.4 & 78.1 & 60.2 & 86.7 & 76.6 & ResNet101 & 88.5 & ViT-B/16 & 63.7 \\
    UniMoS~\cite{unimos}  & A & ResNet50 & 59.5 & 89.4 & 86.9 & 75.2 & 89.6 & 86.8 & 75.4 & 58.4 & 87.2 & 76.9 & 59.5 & 89.7 & 77.9 & ResNet101 & 88.1 & ViT-B/16 & 63.6 \\
    DAMP~\cite{damp}  & C,T & ResNet50 & 59.7 & 88.5 & 86.8 & 76.6 & 88.9 & 87.0 & 76.3 & 59.6 & 87.1 & 77.0 & 61.0 & 89.9 & 78.2 & ResNet101 & 88.4 & ViT-B/16 & 63.2 \\
    CMKD~\cite{zhou2024unsupervised}  & KD & ResNet50 & 65.9 & 86.6 & 87.3 & 74.4 & 87.7 & 85.8 & 75.9 & 64.4 & 87.9 & 79.1 & 67.2 & 90.0 & 79.3 & ResNet101 & 87.0 & ViT-B/16 & 53.9 \\
    DAPrompt~\cite{daprompt} & T & ViT-B/16 & 70.6 & 90.2 & 91.0 & 84.9 & 89.2 & 90.9 & 84.8 & 70.5 & 90.6 & 84.8 & 70.1 & 90.8 & 84.0 & ViT-B/16 & 89.5 & - & - \\
    PDA~\cite{bai2024prompt} & V,T & ViT-B/16 & 73.5 & 91.4 & 91.3 & 86.0 & 91.6 & 91.5 & 86.0 & 73.5 & 91.7 & 86.4 & 73.0 & 92.4 & 85.7 & ViT-B/16 & 89.7 & - & -\\
    ADCLIP~\cite{singha2023ad} & T & ViT-B/16 & 70.9 & 92.5 & 92.1 & 85.4 & 92.4 & 92.5 & 86.7 & 74.3 & 93.0 & 86.9 & 72.6 & 93.8 & 86.1 & ViT-B/16 & 90.7 & - & -\\
    PADCLIP~\cite{lai2023padclip} & FT-R & ViT-B/16 & 76.4 & 90.6 & 90.8 & 86.7 & 92.3 & 92.0 & 86.0 & 74.5 & 91.5 & 86.9 & 79.1 & 93.1 & 86.7 & ViT-B/16 & 90.9 & - & -\\
    UniMoS~\cite{unimos} & A & ViT-B/16 & 74.9 & 94.0 & 92.5 & 86.4 & 94.3 & 92.5 & 86.0 & 73.9 & 93.0 & 86.4 & 74.2 & 94.5 & 86.9 & ViT-B/16 & 90.1 & - & -\\
    DAMP~\cite{damp} & C,T & ViT-B/16 & {\color[HTML]{242021} 75.7} & 94.2 & 92.0 & 86.3 & 94.2 & 91.9 & 86.2 & 76.3 & 92.4 & 86.1 & 75.6 & 94.0 & 87.1 & ViT-B/16 & 90.9 & - & -\\
    Lai \textit{et al.}~\cite{lai2024empowering} & A & ViT-B/16 & 78.2 & 90.4 & 91.0 & 87.5 & 91.9 & 92.3 & 86.7 & 79.7 & 90.9 & 86.4 & 79.4 & 93.5 & 87.3 & ViT-B/16 & 91.7 & - & - \\
    CMKD~\cite{zhou2024unsupervised} & KD & ViT-B/16 & 79.4 & 94.2 & 92.7 & 86.3 & 93.4 & 92.2 & 86.7 & 79.5 & 92.1 & 88.2 & 81.2 & 94.5 & 88.4 & ViT-B/16 & 90.3 & - & - \\\midrule
    \multicolumn{20}{@{}l}{\cellcolor[HTML]{EFEFEF}\textit{Source-free domain adaptation (SFDA) methods}} \\
    DIFO~\cite{tang2024source} & KD & ResNet50 & 62.6 & 87.5 & 87.1 & 79.5 & 87.9 & 87.4 & 78.3 & 63.4 & 88.1 & 80.0 & 63.3 & 87.7 & 79.4 & ResNet101 & 88.8 & - & - \\
    ProDe~\cite{tang2024proxy} & KD & ResNet50 & 64.0 & 90.0 & 88.3 & 81.1 & 90.1 & 88.6 & 79.8 & 65.4 & 89.0 & 80.9 & 65.5 & 90.2 & 81.1 & ResNet101 & 88.7 & - & - \\
    ViLAaD~\cite{tarashima2025vilaad} & KD & ViT-B/32 & 70.1 & 91.6 & 89.9 & 83.2 & 92.0 & 90.0 & 81.0 & 71.7 & 89.9 & 83.3 & 71.3 & 92.3 & 83.9 & ViT-B/32 & 90.5 & - & - \\
    RCL~\cite{chen2024empowering} & KD & ViT-B/32 & 83.1 & 95.7 & 93.1 & 89.2 & 95.3 & 92.6 & 89.2 & 82.3 & 92.9 & 90.0 & 83.2 & 95.5 & 90.2 & ViT-B/32 & 93.2 & - & - \\
    Zhan \textit{et al.}~\cite{zhan2024towards} & KD & ViT-B/16 & 71.1 & 87.1 & 91.3 & 86.3 & 90.9 & 91.6 & 86.6 & 74.1 & 91.8 & 87.6 & 75.0 & 91.8 & 85.4 & - & - & - & - \\
    Co-learn++~\cite{zhang2025source} & KD & ModelZoo$^\dagger$ & 80.0 & 91.2 & 91.8 & 83.4 & 92.7 & 91.3 & 83.4 & 78.9 & 92.0 & 85.5 & 80.6 & 94.7 & 87.1 & ModelZoo$^\dagger$ & 91.1 & - & - \\\bottomrule
    \multicolumn{18}{@{}l}{$^\dagger$ ModelZoo means the results are from a mixture of different pretrained models detailed in \cite{zhang2025source}.}
    \end{tabular}}
\end{table*}

\subsubsection{Dataset-Level Generalization}
\label{sec:experiment_result_1}
In dataset-level generalization, VLMs need to adapt to various tasks with different scenarios, categories and data distributions. \cref{tab:result_1} presents comprehensive results and comparisons over various methods introduced in this survey. Following current research lines, we present dataset-level VLM generalization tasks evaluated on the ImageNet family and other 10 specialized or fine-grained image datasets introduced in \cref{tab:dataset}.  There are mainly three settings that tackle dataset-level generalization. \textbf{(1) Few-shot learning (FSL)} provides few-shot labeled training data for each dataset and each category. Note that FSL on the ImageNet family only trains on ImageNet-1k (-I) and evaluates on all other variants. Some FSL methods split dataset categories into base and novel parts, where the model is only adapted with data from base classes. Harmonic mean~\cite{xian2017zero} is adopted to assess the general model performance across both base and novel classes. \textbf{(2) Domain generalization (DG)} assumes no available data on the evaluated datasets, where the model is only trained on ImageNet and expected to generalize to unseen tasks. \textbf{(3) Test-time adaptation (TTA)} directly generalizes and evaluates pretrained VLMs on target datasets on-the-fly without the training process. 

From the results we can conclude the following factors that affect the classification performance. \textbf{(1) Backbones.} The majority of methods base on the vision transformer (ViT) family~\cite{dosovitskiy2020image} and ResNet family~\cite{he2016deep}, where ViT-based method exhibit a significant accuracy gain of $\sim$15\%. Such phenomenon indicates that the generalization ability of VLMs is largely related to that of vision encoders. \textbf{(2) Transfer settings.} With labeled samples from the target dataset, FSL methods exhibit the best overall performance on few-shot datasets. The overall performances of DG and TTA methods are comparable. Specifically, on hard OOD datasets with large distribution gaps, e.g., ImageNet-A, FGVC, EuroSAT, TTA methods exhibit superior performances. The reason is that TTA methods can access these hard samples and make quick adjustments to mitigate the gap. On more general tasks with in-distribution data, DG methods perform better due to sufficient tuning on source data with similar distributions. \textbf{(3) Method designs.} Under the same setting and backbone, different method types contribute to the subtle performance fluctuations. For example, on FSL tasks, prompt-tuning methods perform generally better at the expenses of higher computation overheads. On the contrary, on TTA tasks, prompt-tuning methods deteriorate due to insufficient training and the absence of ground truths. Cache models are better at capturing the representative features of streamlined data flows, making them suitable for TTA tasks. Distillation-based methods (KD methods) surpass other competitors by introducing external knowledge from larger teacher models (ViT-L/14 in \cref{tab:result_1}).

\subsubsection{Domain-Level Generalization}
\label{sec:experiment_result_2}
On domain-level DG tasks, each dataset is composed of several distinct \textit{domains} with the same label space but different data distribution. \cref{tab:result_2} presents evaluation results based on the five domain-level DG datasets (OH is short for OfficeHome, DN is short for DomainNet, TI is short for TerraIncognita) established in DomainBed~\cite{gulrajani2020search}.  The methods follow leave-one-out protocol that generalizes to each target domain by tuning on all other domains. The reported accuracies are averaged over all tested target domains. It can be observed that fine-tuning-based (FT) methods~\cite{clipood,mao2024context} perform generally better than other methods, especially on harder datasets like TerraIncognita. The reason is that pretrained VLMs cannot overcome significant domain gaps without full fine-tuning~\cite{lai2023padclip}. The adapter-based methods achieve slightly better results on easier tasks where CLIP's zero-shot accuracies are high, but fall far behind FT methods on challenging tasks, negatively affecting their overall performances. However, the absolute accuracies of FT methods on large distribution shifts (around 60\%) still leave significant room for improvements, highlighting the challenges of generalizing pretrained VLMs to unseen domains that deviate far from the training data.

\subsubsection{Domain-Wise Transfer}
\label{sec:experiment_result_3}
While domain-level generalization trains on the mixture of source domains, domain-wise transfer performs finer adaptation from exactly one source domain to one target domain. As shown in \cref{tab:result_3}, for dataset OfficeHome with domains \textit{Art (A)}, \textit{Clipart (C)}, \textit{Product (P)} and \textit{RealWorld (R)}, knowledge transfer is performed on each source-target pair, leading to 12 subtasks in total. DomainNet includes 6 domains and 30 cross-domain tasks, and their means are reported due to space limits. Please refer to the origin paper for full domain-wise results. On VisDA there is only one synthetic-to-real adaptation task and the mean-class accuracies are reported following previous works. Note that the CLIP's zero-shot results on tasks with the \textit{same target domain} are \textit{identical} since the model is not trained on the source domain. 

From \cref{tab:result_3} we can make a similar observation that the backbone of CLIP's vision encoder is the main factor that affects the performance. However, the accuracy gap brought by backbone is narrowing with the advances in research works, especially on easier tasks (e.g., AP, CP, RP on OfficeHome and VisDA). Unlike dataset-level generalizations, the differences in method design do not bring significant changes in performance. It is worth noting that CLIP-based UDA methods significantly surpass single-modality baselines, mainly due to the strong zero-shot ability of pretrained CLIP - even CLIP's zero-shot accuracies easily surpass that of various single-modality methods. However, current CLIP-based UDA methods overly rely on pseudo labels generated by CLIP, leading to \textit{similar} accuracies among tasks with the same target domain. This phenomenon indicates that the source knowledge are still under-explored, thus hardly affecting the transfer performance. Such observation also holds true for SFDA tasks.
\begin{table*}[!t]
    \large 
    \caption{Summarization of representative multimodal large language models (MLLM). The \textbf{F} column represents the MLLM family.}
    \centering
    \label{tab:mllm}
    \vspace{-8pt}
    \resizebox{\linewidth}{!}{
    \setlength{\tabcolsep}{4.9pt}
    \begin{tabular}{@{}cm{1.8cm}m{1.4cm}m{8.65cm}m{2.2cm}m{1.92cm}m{1.98cm}m{12.9cm}@{}}
    \toprule
    \textbf{F} & \textbf{Model} & \textbf{Release date} & \textbf{Training datasets} & \textbf{Vision encoder} & \textbf{VL Adapter} & \textbf{LLM} & \textbf{Brief description} \\ \midrule
     & DeepSeek-VL~\cite{lu2024deepseek} & 2024.3 & Image-text, image caption, table \& charts, web code, scene text OCR, document OCR, text corpus, in-house data, shared gpt4v datasets.  & SigLIP~\cite{zhai2023sigmoid}, SAM-B \cite{kirillov2023segment}, ViTDet~\cite{li2022exploring} & two-layer MLP & DeepSeek LLM~\cite{bi2024deepseek} & Introduces hybrid vision encoder to efficiently handle high resolution inputs. Proposes to preserve the language ability of VLMs by dynamically adjust the ration between vision and text modaliy data. \\[20pt]
     \multirow{4}{*}{\rotatebox{90}{\textbf{DeepSeek}}} & DeepSeek-VL2~\cite{wu2024deepseek} & 2024.12 & Image-text, image captioning, OCR, visual QA, visual grounding, conversations, reasoning, logic, mathematics, textbook, academic questions, code generation, etc. & SigLIP-SO400M \cite{zhai2023sigmoid} & two-layer MLP & DeepSeek MoE~\cite{dai2024deepseekmoe} & Introduces tiling strategy for processing high-resolution images with different aspect ratios. Replaces the LLM with DeepSeek-MoE with the Multi-head Latent Attention mechanism. \\[20pt]
     & DeepSeek Janus~\cite{wu2025janus} & 2024.10 & Image-text paired captions~\cite{chen2024sharegpt4v}, images, text, table \& charts, visual generation data.  & SigLIP-Large~\cite{zhai2023sigmoid} & task-specific adapters & DeepSeek LLM~\cite{bi2024deepseek} & Decouples the visual encoding process into separate encoders to handle multimodal understanding and generation, respectively. \\[20pt]
     & DeepSeek Janus-Pro~\cite{chen2025janus} & 2025.1 & Image caption, table, chart,  document understanding data, aesthetic data, and datasets from DeepSeek-VL2. & SigLIP-Large~\cite{zhai2023sigmoid} & task-specific adapters & DeepSeek LLM~\cite{bi2024deepseek} & Improves DeepSeek Janus with extended training data and model scale. Improves the train strategy by focusing on the training of VL-adapter and image head, as well as image-text data for  pretraining. \\[20pt]\midrule
     & Qwen-VL~\cite{bai2023qwenvlversatilevisionlanguagemodel} & 2023.10 & LAION-5B~\cite{schuhmann2022laion},  LAION-COCO~\cite{schuhmann2022laionb}, DataComp~\cite{gadre2023datacomp},  Coyo~\cite{kakaobrain2022coyo700m}, CC12M~\cite{changpinyo2021conceptual},  CC3M~\cite{sharma2018conceptual}, SBU~\cite{ordonez2011im2text}, COCO  Caption~\cite{chen2015microsoft} and VL annotation data.  & OpenCLIP ViT-bigG~\cite{ilharco_2021_5143773} & single-layer cross-attn. & Qwen 7B~\cite{Qwen2023} & Qwen-VL introduces a vision encoder, VL adapter and a 3-stage training pipeline to endow Qwen-LM with various multimodal abilities. An addition set of dialogue data incorporates localization and multi-image comprehension abilities into the model. \\[25pt]
     \multirow{2}{*}{\rotatebox{90}{\textbf{Qwen}}} & Qwen2-VL~\cite{wang2024qwen2} & 2024.10 & Image-text pairs, OCR data, interleaved image-text articles, visual QA datasets, video dialogues, image knowledge datasets, etc.  & Modified DFN's ViT~\cite{fang2023data} & single-layer cross-attn. & Qwen2 \cite{team2024qwen2} & Proposes Naive Dynamic Resolution mechanism to process dynamic image resolutions, and Multimodal Rotary Position Embedding (MRPE) to fuse positional information across modalities.\\[20pt] 
     & Qwen2.5-VL~\cite{bai2025qwen2} & 2025.2 & Image caption, OCR, pure text, interleaved data, vision QA, video grounding, agent, long video, document, etc.  & Modified   ViT & two-layer MLP & Qwen2.5 \cite{qwen2025qwen25technicalreport} & Building on Qwen-VL2, proposes MRPE aligned to absolute time, native dynamic resolution and redesigned ViT to achieve strong generalization ability across domains without specific fine-tuning. \\[20pt]\midrule
     & BLIP~\cite{li2022blip} & 2022.2 & 129M  images from COCO, Visual Genome \cite{krishna2017visual}, Conceptual Captions \cite{changpinyo2021conceptual}, CC 12M \cite{changpinyo2021conceptual}, SBU captions \cite{ordonez2011im2text}, LAION400M \cite{schuhmann2021laion}. & ImageNet ViT-B and ViT-L & None & BERT-base & A multimodal mixture of encoder-decoder structure, composed of aligned unimodal encoders, image-grounded text encoder, and image-grounded text decoder to generate answers. \\[20pt]
     \multirow{4}{*}{\rotatebox{90}{\textbf{BLIP}}} & BLIP2~\cite{li2023blip} & 2023.1  & The same as BLIP. & CLIP ViT-L, EVA-CLIP ViT-g~\cite{fang2023eva} & Q-Former & OPT~\cite{zhang2022opt}, FlanT5~\cite{chung2024scaling} & Integrates a lightweight Query Transformer (Q-Transformer) to extract vision features that mitigate the modality gap between frozen image encoder and LLM. \\[20pt]
     & Instruct-BLIP~\cite{dai2023instructblipgeneralpurposevisionlanguagemodels} & 2023.6 & Datasets including image captioning, visual reasoning, visual QA, image question generation, video QA, image classification, etc.  & EVA-CLIP ViT-g/14~\cite{fang2023eva} & Q-Former with instructions & FlanT5~\cite{chung2024scaling}, Vicuna \cite{chiang2023vicuna} & Introduces instructions to Q-Former of BLIP2 to extract instruction-aware visual features. The training datasets are extended to a wider range of multimodal tasks in instruction tuning format. \\[20pt]
     & xGen-MM~\cite{xue2024xgen} & 2024.8 & Interleaved dataset mixture, including MINT-1T~\cite{awadalla2024mint}, OBELICS~\cite{laurenccon2024matters}, BLIP3-KALE, BLIP3-OCR-200M, BLIP3-GROUNDING-50M, and other public datasets. & CLIP-DFN \cite{fang2023data}, SigLIP~\cite{zhai2023sigmoid} & perceiver resampler~\cite{alayrac2022flamingo} & Phi3-mini~\cite{abdin2024phi} & Scales up BLIP2 with an ensemble of multimodal interleaved datasets, and replaces Q-Former with a scalable vision token sampler to down sample the encoded any-resolution vision tokens. \\[20pt]\midrule
     & LLaVA \cite{liu2023visual} & 2023.4 & COCO images~\cite{lin2014microsoft}, filtered CC3M~\cite{sharma2018conceptual}, transformed to instruction data using GPT-4. & CLIP ViT-L/14 & single linear layer & Vicuna \cite{chiang2023vicuna} & Generates instruction-following data from multimodal data using GPT-4, then trains a MLLM that connects an image encoder and LLM for mutlimodal chat, science QA, etc. \\[20pt]
     & LLaVA-1.5~\cite{liu2024improved} & 2023.10 & The same as LLaVA, plus open-knowledge VQA and OCR data. & CLIP ViT-L 336px & two-layer MLP & Vicuna \cite{chiang2023vicuna} & Improves LLaVA by replaceing the VL adapter with a two-layer MLP, improved QA data format, and grid-split the images to up-scale to high resolustion visual perception. \\[20pt]
     \multirow{2}{*}{\rotatebox{90}{\textbf{LLaVA}}} & LLaVA-plus~\cite{liu2024llava} & 2023.11 & LLaVA data and curated tool-use instruction data.  & \multicolumn{3}{c}{Based on pretrained MLLM weights.} & A generalizable agent that uses a large and diverse set of  external tools according to users' needs, trained in an end-to-end fashion. \\[15pt]
     & LLaVA-OneVision \cite{li2024llava} & 2024.10 &  Re-captioned description data, document / OCR data, Chinese and language data, visual instruction data, etc.  & SigLIP~\cite{zhai2023sigmoid} & two-layer MLP & Qwen2 \cite{team2024qwen2} & A family of MLLM to handle single-, multi-image and videos with transfer learning ability across modalities and domains, supported by an AnyRes strategy to handle different resolutions. \\[20pt]
     & Dynamic-LLaVA \cite{huang2024dynamic} & 2024.12 &  The same as LLaVA-1.5. & \multicolumn{3}{c}{Based on pretrained MLLM weights.} & A dynamic vision-language context sparsification framework to save memory consumption, GPU memory, and inference time, by reducing redundancy in prefill and decoding stages. \\[20pt]\midrule
    \multirow{2}{*}{\rotatebox{90}{\textbf{PaliGemma}}} & Pali-Gemma \cite{beyer2024paligemma} & 2024.7 & Image   captioning, visual QA, image segmentation, video, etc. Please refer to the paper for full details. & SigLIP-SO400M \cite{zhai2023sigmoid} & single linear layer & Gemma-2B~\cite{team2024gemma} & A family of MLLM following the PaLI family~\cite{chen2023pali,chen2023palib} to handle images of various resolutions, and can be transferred to domain-specific tasks including remote-sensing and segmentation. \\[20pt]
     & Pali-Gemma 2~\cite{steiner2024paligemma} & 2024.12 & Data in PaliGemma, plus text detection, table recognition, molecular structure, optical music score, long caption generation, spatial reasoning, radiography report generation. & SigLIP-SO400M \cite{zhai2023sigmoid} & single linear layer & Gemma 2~\cite{team2024gemma} & Enhances PaliGemma with improved LLM and a wider range of transfer tasks including OCR data, fine-grained captioning and radiography report generation. \\[20pt]\midrule
    \multirow{2}{*}{\rotatebox{90}{\textbf{Apple MM}}} & MM1~\cite{mckinzie2024mm1} & 2024.4  & Mixture of 45\% captioned images, 45\% interleaved image-text document, 10\% text.  & ViT trained on DFN~\cite{fang2023data} & C-Abstractor \cite{cha2024honeybee} & Private LLM & Investigates several factors when building a MLLM, e.g., data composition, image encoder, VL adapter, pretrain loss. Builds a family of MLLM with enhanced in-context, multi-image, few-shot abilities. \\[20pt]
     & MM1.5 \cite{zhang2024mm1} & 2024.9 & Data in MM1, plus OCR data, and newly introduced supervised fine tuning data with optimal ratio.  & ViT-H   trained on DFN~\cite{fang2023data} & C-Abstractor \cite{cha2024honeybee} & Private   LLM & Improves   MM1 from the perspective of data composition by exploring optimal data   mixture ratios and curation. Two specialized variants, MM1.5-Video and   MM1.5-UI are tailored for video and mobile UI understanding. \\[20pt]\midrule
    \multirow{2}{*}{\rotatebox{90}{\textbf{MiniGPT}}} & MiniGPT-4~\cite{zhu2023minigpt} & 2023.10 & Conceptual Captions~\cite{changpinyo2021conceptual}, CC 12M~\cite{changpinyo2021conceptual}, SBU~\cite{ordonez2011im2text}, LAION400M~\cite{schuhmann2021laion}, and a newly curated dataset. & EVA-CLIP ViT-g~\cite{fang2023eva} as in \cite{li2023blip} & single linear layer & Vicuna \cite{chiang2023vicuna} & Demonstrates that by connecting a pretrained vision encoder and LLM with one projection layer, the resultant MLLM can achieve similar multimodal abilities as in GPT4. \\[20pt]
     & MiniGPT-v2~\cite{chen2023minigpt} & 2023.11 &CC3M~\cite{sharma2018conceptual},  SBU~\cite{ordonez2011im2text}, LAION400M~\cite{schuhmann2021laion}, vision QA   datasets, LLaVA instruction data~\cite{liu2023visual}, Flickr30k   entities~\cite{plummer2015flickr30k}, KOSMOS-2~\cite{peng2023kosmos} & EVA-CLIP ViT-g \cite{fang2023eva} & single linear layer & LLaMA2-chat~\cite{touvron2023llama} & Builds a unified interface for various vision-language tasks, i.e., a MLLM, on   pretrained LLM and vision encoder using instruction-format data.\\[20pt]\midrule 
    \multirow{2}{*}{\rotatebox{90}{\textbf{ByteDance}}} & Seed1.5-VL~\cite{guo2025seed1} & 2025.6 & Image-text, OCR, visual grounding, STEM, video, GUI, 3D Spatial, long Chain-of-Thought data, and RLHF.  & private Seed-ViT  & two-layer MLP & private Seed1.5-LLM & Latest 20B MLLM obtained with high-quality synthetic vision-language data and hybrid training infrastructure to achieve leading multimodal abilities with reduced expenses. \\[20pt]
    & BAGEL \cite{deng2025emerging} & 2025.6 & Pure text, image-text, vision-text interleaved data, video, reasoning-augmented data (e.g., image manipulation, conceptual edits). & SigLIP2-so400m/14 \cite{tschannen2025siglip} & two-layer MLP & Qwen2.5 LLM \cite{qwen2025qwen25technicalreport} & An unified 7B MLLM supporting general understanding and generation tasks across modalities (e.g., free-form visual manipulation) with mixure-of-Transformer structure. \\[15pt] \bottomrule 
    \end{tabular}}
\end{table*}
\section{Multimodal Large Models}
\label{sec:mllm}
Recent years have witnessed the revolutionary breakthroughs and astonishing capabilities of large language models (LLMs), exemplified by the GPT family~\cite{brown2020language,achiam2023gpt}. As the language branch plays an important part in VLMs for encoding and aligning textual information, one intuitive question is that, \textit{is it possible to enhance VLMs by integrating the general-purpose LLMs?} Research efforts have proven the viability of such approach. Furthermore, the resultant LLMs enhanced by pretrained VLMs show strong generalization abilities to a wide range of vision-language tasks, e.g., visual question answering (QA), segmentation, captioning, etc. These large models are summarized as multimodal large language models (MLLMs), due to their capabilities in handling vision and text in various forms. This section first introduces typical structures and training strategies for general MLLMs, then conduct a more detailed review on representative MLLMs.

\begin{figure}[!t]
    \centering
    \includegraphics[width=0.49\textwidth]{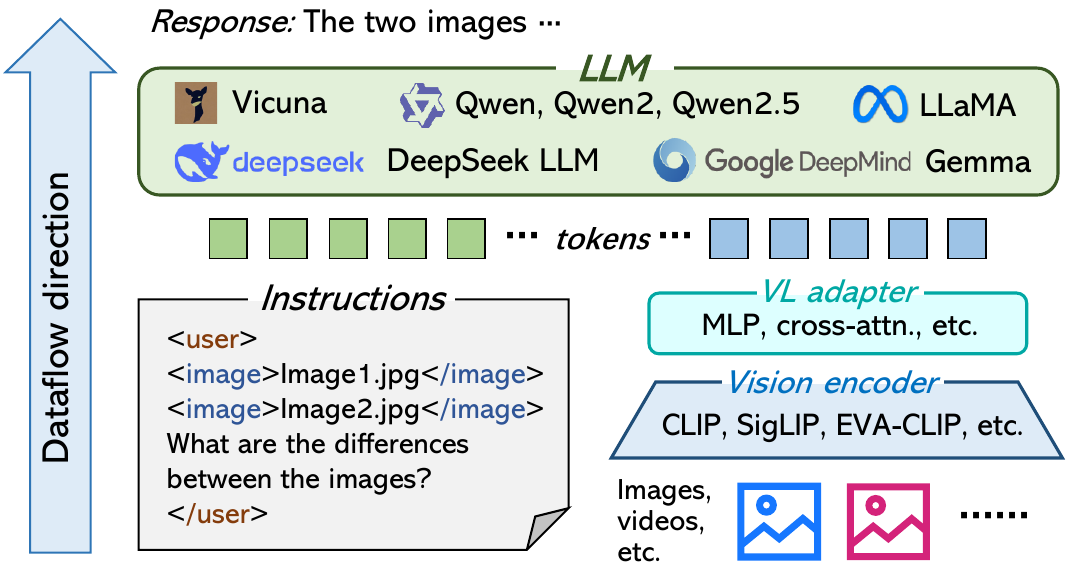}
    \vspace{-4pt}
    \caption{General structure of MLLMs, composing of a pretrained vision encoder, an LLM and a VL-adapter to connect the two. The inputs include vision-modality materials (images, videos) and textual instructions.}
    \label{fig:mllm}
    \vspace{-4pt}
\end{figure}

\subsection{MLLM Structure}
\label{sec:mllm_structure}
Current MLLMs are mostly obtained by integrating a vision encoder with an LLM, as shown in \cref{fig:mllm}. The vision encoder perceives various visual information, including videos, images of different resolutions, and transforms them into high-dimensional features. A lightweight vision-language adapter (VL adapter) aligns the dimension and distribution between the vision and text representation space. The resultant vision tokens are concatenated with text tokens of corresponding instruction data, which are then sent to an LLM to generate the desired outputs. In practice, the tokens may be concatenation of multilingual texts, images of different scales, videos, and potentially more modalities encoded by other encoders~\cite{li2024baichuan,li2025baichuan}. 

To mitigate modality gap~\cite{liang2022mind} between vision and text tokens, current MLLMs~\cite{zhu2023minigpt,li2023blip,liu2023visual,bai2023qwenvlversatilevisionlanguagemodel} typically adopt the vision branch of pretrained VLMs~\cite{clip,zhai2023sigmoid,fang2023eva,fang2023data} as the vision encoder. The choice of LLM varies for different research teams, where individual researchers usually adopt public LLMs~\cite{chiang2023vicuna,chung2024scaling}, while AI institutions prefer their own LLMs trained with in-house data~\cite{li2024baichuan,mckinzie2024mm1,dai2024deepseekmoe}. The adapters are usually one- or two-layer MLP.

Training data and strategy is the key factor that decides model performances. Introduced in \cite{liu2023visual}, instruction tuning~\cite{ouyang2022training} has been widely adopted for tuning MLLMs to follow human instructions for various tasks. Multimodal datasets, e.g., captioning~\cite{changpinyo2021conceptual,ordonez2011im2text}, visual QA~\cite{hudson2019gqa,goyal2017making,krishna2017visual}, image-text pairs~\cite{schuhmann2021laion,schuhmann2022laion,changpinyo2021conceptual}, are then transformed to instruction-following data using automatic pipelines like GPT~\cite{achiam2023gpt}. Every of the three components in the MLLM is trainable. Considering their parameter count and functionalities are fundamentally different, current methods adopt multi-stage training paradigm to adjust different components with different data and tuning strategy. Finer discussions on the training data, structure, training strategy of different MLLMs are provided in the following section.

\subsection{General-Purpose MLLMs}
\label{sec:mllm_discussion}
\cref{tab:mllm} introduces recent advances (2022-2025.6) of MLLMs. Note that the \textbf{Training datasets} column are high-level summarization of the vast training data. For more details please refer to the original papers.

\subsubsection{Research-Oriented MLLMs}
\label{sec:mllm_discussion_research}
\textbf{BLIP family} are one of the earliest efforts in building MLLMs. Similar to its ancestor CLIP~\cite{clip}, BLIP builds on web-crawled image-text data but with denoising. Unlike the two-towel structure of CLIP, BLIP introduces mixture of encoder-decoder trained with multiple cross-modality alignment losses. BLIP is based on single-modality vision encoders (ImageNet-pretrained ViT) and small-scale language decoders, which is more like a VLM rather than a MLLM. One year later, the same research team proposes BLIP2~\cite{li2023blip}, a MLLM following the structure in \cref{fig:mllm} using EVA-CLIP~\cite{fang2023eva} vision encoder and OPT~\cite{zhang2022opt}, FlanT5~\cite{chung2024scaling} as LLM. The VL-adapter is a newly-proposed Query Transformer (Q-Former) to mitigate modality gap. BLIP2 adopts a two-stage training paradigm to endow Q-Former with multimodal representation abilities, while keeping the pretrained vision encoder and LLM frozen. Based on BLIP2, InstructBLIP~\cite{dai2023instructblipgeneralpurposevisionlanguagemodels}  further introduces instruction-tuning for the training of Q-Former to obtain a general vision-language task solver. Recently, a family of MLLM termed xGen-MM~\cite{xue2024xgen}, also known as BLIP-3, have been released. With high-quality interleaved datasets and improved VL-adapter~\cite{alayrac2022flamingo}, BLIP-3 achieves competitive performances on various vision-language tasks.

\textbf{LLaVA family} are another group of instruction-tuned MLLMs built upon public VLMs and LLMs. Aiming to conduct visual instruction tuning, Liu \textit{et al.}~\cite{liu2023visual} propose to transform available image-text data into instruction-following data by prompting language-only LLMs, e.g., GPT4~\cite{achiam2023gpt}. Based on instruction data, LLaVA~\cite{liu2023visual} is obtained by tuning an LLM Vicuna~\cite{chiang2023vicuna} and a single projection layer in a two-step paradigm. The vision encoder from pretrained CLIP remains frozen all the time. With systematic exploration on the training principles of LLaVA, including data composition, data scale, choices in MLLM components, a more powerful variant LLaVA-1.5~\cite{liu2024improved} is obtained. Building on the success of LLaVA family, follow-up researches have explored multimodal agents~\cite{liu2024llava} that operate on various tools to complete uses' complex requirements, or try to reduce computation resource consumption when building MLLMs~\cite{huang2024dynamic}. LLaVA-OneVision~\cite{li2024llava} summarizes the extensions and advances of the LLaVA-NeXT series achieved by handling high-resolution images, improving data quality and LLM capabilities. 

\textbf{MiniGPT family} are early efforts to build MLLMs that reach the capabilities of private GPT4, using public base models and datasets. By aligning the vision encoder and Q-Former in BLIP2~\cite{li2023blip} with the Vicuna~\cite{chiang2023vicuna} LLM using instruction-data-tuned VL adapter, MiniGPT-4~\cite{zhu2023minigpt} achieves comparable abilities with GPT4. Similar observations are obtained in MiniGPT-v2~\cite{chen2023minigpt} with different base models. Differently, MiniGPT-v2 also updates parameters in the LLM with a coarse-to-fine three-step training scheme.

\subsubsection{Commercial MLLMs}
\label{sec:mllm_discussion_commercial}
Although commercial and research-oriented MLLMs share similar structures, the abundant high-quality in-house training data and computation resources in large AI institutions allow more in-depth tuning of MLLM parameters, resulting in more powerful and practical commercial MLLMs.

\textbf{Qwen family} are based on Alibaba's Qwen-LLM series~\cite{Qwen2023,team2024qwen2,qwen2025qwen25technicalreport}. Qwen-VL~\cite{bai2023qwenvlversatilevisionlanguagemodel} is trained with three stages: general pretraining of \textit{vision encoder and VL adapter} with weakly-supervised image-text pairs, multi-task pretraining of \textit{the whole model} with high-quality interleaved mutli-task vision-language data, and supervised fine-tuning of \textit{LLM and VL adapter} with instruction-following data. Instead of tuning a small part of MLLM as in research-oriented MLLMs, we can observe that commercial MLLMs generally tune the whole model progressively for better task generalization. One critical challenge in building MLLMs is how to handle various input image resolutions. Qwen2-VL~\cite{wang2024qwen2} tackles this by introducing dynamic resolution support~\cite{dehghani2023patch} and Multimodal Rotary Position Embedding (M-RoPE) to encode position information of multimodal inputs. Qwen2.5-VL~\cite{bai2025qwen2} further advances the MLLM on various modalities, e.g., video, by considering the information in spatial and temporal dimensions.

\textbf{DeepSeek family} are famous for their superior performance and low training costs. DeepSeek-VL~\cite{lu2024deepseek} focuses on the preservation of language abilities when building MLLMs. To achieve this, a local-to-global training strategy is applied, where the VL-adapter is first warmed-up, followed by joint pretraining of LLM and VL-adapter. Finally, the whole model is fine-tuned with instruction-following data. Specifically, the ratio of language and multimodal data in step 2 is 7:3 to alleviate the loss of language abilities. DeepSeek-VL2~\cite{wu2024deepseek} replaces the DeepSeek-LLM~\cite{bi2024deepseek} with DeepSeek-MoE~\cite{dai2024deepseekmoe} for more efficient inference. A dynamic tiling vision encoding strategy is proposed to handle high-solution image inputs, which divides large images into tiles. DeepSeek-Janus~\cite{wu2025janus} is a unified autoregressive framework with mutlimodal understanding and generation ability. Slightly different from \cref{fig:mllm}, Janus introduces separate understanding/generation encoders and decoders to handle both tasks in isolation. The training procedure is similar to DeepSeek-VL, except the generation decoder is also trainable in step 1, and only the understanding encoder is trainable in step 3. The improved version Janus-Pro~\cite{chen2025janus} extends the data and model scale. The training strategy is improved by lengthening train step 1 on pure images, and drop these images in step 2 to focus on image-text data. 

\textbf{PaliGemma family} are open-source variants of the Gemini series~\cite{team2023gemini,team2024gemini}, built on SigLIP~\cite{zhai2023sigmoid} and public Gemma~\cite{team2024gemma} LLM. PaliGemma follows a pretrain - transfer strategy. The whole model is first pretrained on extensive multimodal tasks and finer trained with increased image resolutions. The obtained base model is then fine-tuned to transfer towards specific goals. 

\textbf{MM family} are private MLLMs developed by Apple. MM focus on exploring data-centric training strategies and reveal several properties when tuning MLLM with different data composition in different stages. For example, MM1.5~\cite{zhang2024mm1} adopts a three-step training recipe: (1) Large-scale pretraining on low-resolution image-text pairs and text-only data. (2) Continual pretraining with high-resolution OCR data. (3) Supervised fine-tuning with mixture of data. In each stage, the impacts of data mixture ratios and each data type are  explored in detail.

\textbf{Seed family} developed by ByteDance are among the latest advances in MLLM. SeedVL-1.5~\cite{guo2025seed1} is built with pretraining and post-training steps. In pretraining, the VL adapter is first warmed-up, followed by full training of the whole model with carefully decided tasks and data mixture ratio. In post-training, the MLLM is equipped with instruction-following and understanding abilities by finely-constructed supervised fine-tuning data, e.g., chain-of-thought data, and boosted by reinforcement learning with human feedback~\cite{ziegler2019fine} or verifiable rewards~\cite{lambert2024t}.

\subsection{Discussions}
\label{sec:mllm_summary}
The development of general MLLMs is a research focus and heated topic in the community, with new generations of MLLM being released monthly. As an extension of VLM, MLLM obtains strong generalization ability across various modalities and language-image tasks by connecting language-aligned image encoders with powerful LLMs. They generally follow the structure in \cref{sec:mllm_structure} with progressively improved base models and training corpus. A part of popular or representative MLLMs are introduced in this survey, categorized into research (\cref{sec:mllm_discussion_research}) or commercial~\cref{sec:mllm_discussion_commercial} purpose MLLMs. It is expected that more advanced MLLMs with more general abilities, e.g., multimodal agents, be released in the future.

\section{Future Directions}
\label{sec:future}

The research on the vision-language field, integrating with advances in large-scale training, is a promising direction in developing modern intelligent systems. The goal is to build general-purpose VLMs generalizable to any data distribution and multimodal tasks. While current models have shown astonishing capabilities in certain tasks, they are still limited by out-of-distribution data and tasks. For example, \textit{none} of the cutting-edge models can solve a most recent coding benchmark~\cite{zheng2025livecodebench}. Therefore, there is still plenty of room for improvements. Some promising directions are discussed as following.

\textbf{(1) Generalize under black-box setting.} Most of the current advances aiming to transfer a general VLM to downstream tasks are based on fully open-source frameworks like CLIP~\cite{clip}. However, the most advanced achievements are generally private commercial services accessed by API, e.g., methods introduced in \cref{sec:mllm_discussion_commercial}. The pretrained models in data-sensitive fields, e.g., medical, military, may also constrain the accessibility of their structures and weights. The goal is to build a target model by transferring the knowledge behind the API. Current attempts~\cite{addepalli2024leveraging,park2025zip} rely on model distillation, but their assumptions on label distribution, accessibility of final-layer features, etc., limit their practicability. A general VLM transfer technique without accessing the base model's weights and structures is needed.

\textbf{(2) Generalize efficiently.} Although current VLM generalization methods adopt various parameter-efficient fine-tuning (PEFT) strategies, e.g., prompt tuning in \cref{sec:prompt}, adapters in \cref{sec:feature_refine_adapter}, their efficiencies are limited to the parameter dimension, but neglecting other factors like memory, GPU consumption. For example, the training resources required by typical prompt tuning in \cite{coop} scales with increasing category numbers, and requires computing gradient on all VLM parameters. More dimensions of transfer efficiency need to be considered. One recent attempt~\cite{hao2024quantized} adopts quantification to achieve generalization with memory efficiency. In order to deploy and adapt more powerful VLMs on resource-constrained scenarios like edge devices, researches on a larger spectrum of generalization efficiency are desirable.

\textbf{(3) General multimodal agents.} Agents are general helpers that aid the user to accomplish goals by operating on various tools, e.g., coding~\cite{zhang2024codeagent}. With general multimodal ability, agents are expected to handle a wider range of tasks. There have been pioneering studies on multimodal agents~\cite{xie2024large}, but they often rely on pretrained LLMs, whose generalizability across modalities, novel tasks and domains remains underexplored. The daily tasks in human society span across various domains and involve substantial operations. Trained and evaluated with limited data, the agents' behavior in open real world cannot be well assessed, and their lack of true generalizability may lead to undesirable performances. Therefore, investigating and improving the generalizability of large-model-driven multimodal agents is a promising research direction. This can be further extended to applications on embodied intelligence.

\textbf{(4) Native vision-language structure.} Current MLLMs are built by connecting pretrained LLMs with language-aligned vision encoders. As the number and complexity of input modalities increase, the demands for the encoders - vision, audio ones, and probably more - are making the overall structure overly complicated and less scalable. The computation limitations posed by the attention mechanism also hinder the models' generalization to the infinite open world. Is it possible to construct an all-in-one multimodal model from scratch, which supports any modality natively? 

\textbf{(5) Generalizability of vision-language training data.} When building general MLLMs, the quality and scale of multimodal training data play a dominant role. However, it is unclear that whether all the data contribute to the generalization of the target model. There have been researches~\cite{niu2023towards,zhu2023prompt} showing that training with domain and modality-aligned data can improve adaptation performance. Extensive researches~\cite{mckinzie2024mm1,zhang2024mm1,chen2025janus} have also investigated the optimal mixture proportion and composition of vision-language data to achieve a balance between model performance and data efficiency. In the era of Internet, the quality of web-crawled data is not guaranteed, and polluted data may even introduce negative effects during training. Therefore, the community is in need of an automatic framework that determine the quality and usefulness of raw multimodal data given target tasks.

\section{Conclusion}
This paper systematically reviews recent advances in generalizing and adapting vision-language models (VLMs) to novel tasks and domains. Based on the mostly adopted VLM structures, this survey divides the generalization methods into prompt-based, parameter-based and feature-based methods, according to the adapted model component. The reviewed methods are introduced based on the traditional transfer learning settings, revealing how these transfer problems are being solved in the era of vision-language studies. Thorough performance comparisons among reviewed methods on standard generalization benchmarks are provided. Following advances of large-scale training, this survey also includes most up-to-date and representative multimodal large language models (MLLMs), which enhances the generalizability of vision-language systems with powerful large language models. Finally, current challenges and future directions of vision-language researches are discussed to inspire further advancements of the community.

%
\IEEEpeerreviewmaketitle

\ifCLASSOPTIONcaptionsoff
  \newpage
\fi



\bibliographystyle{IEEEtran}
\bibliography{ref}
\end{document}